\documentclass[10pt]{article}
\usepackage[a4paper, total={5.5in, 8.5in}]{geometry}
\usepackage{times}
\usepackage{epsfig}

\usepackage[utf8]{inputenc}
\usepackage{graphicx,verbatimbox}
\usepackage{tcolorbox}

\usepackage{rotating}
\usepackage{palatino}

\usepackage{tikz}
\usepackage{comment}
\usepackage{amsmath,amssymb} %

\usepackage{booktabs} %
\usepackage{xcolor}
\usepackage{color,colortbl}
\usepackage{pifont}
\usepackage[colorlinks=true,linkcolor=blue, citecolor=cyan]{hyperref}%

\usepackage{multirow}
\usepackage[font=footnotesize]{caption}
\usepackage{placeins}
   
\usepackage{xspace}

\usepackage[accsupp]{axessibility}  %

\usepackage{booktabs}
\usepackage{multirow}

\usepackage{amsmath, bm}
\usepackage[linesnumbered,ruled]{algorithm2e}
\usepackage[font=small,labelfont=bf]{caption}
\usepackage{subcaption}
\usepackage{multirow}
\usepackage{graphicx}
\usepackage{epsfig}
\usepackage{tabularx}
\usepackage{mathtools}
\usepackage{tabulary}
\usepackage{booktabs}       
\usepackage{nicefrac}       
\usepackage{microtype}      
\usepackage{booktabs}
\usepackage{pifont}

\definecolor{Goldenrod}{RGB}{245,245,220}

\usepackage[numbers,sort&compress]{natbib}

\title{Efficient Vision-Language Pretraining with Visual Concepts and Hierarchical Alignment} 
\author{
\begin{minipage}{\linewidth}
\begin{center}
\normalsize Mustafa Shukor$^{\diamond}$\thanks{Corresponding author: mustafa.shukor@sorbonne-universite.fr} \hspace{0.23cm} Guillaume Couairon$^{\diamond,\dagger}$ \hspace{0.23cm} Matthieu Cord$^{\diamond, \ddagger}$ \\[0.5cm] 
\scalebox{1.}{$^\diamond$Sorbonne University\hspace{0.5cm} $^\dagger$Meta AI\hspace{0.5cm} $^\ddagger$Valeo.ai}\\
\end{center}
\end{minipage}
}

\date{~}

\begin{document}

\maketitle

\begin{abstract}
Vision and Language Pretraining has become the prevalent approach for tackling multimodal downstream tasks. 
The current trend is to move towards ever larger models and pretraining datasets. This computational headlong rush does not seem reasonable in the long term to move toward sustainable solutions, and \textit{de facto} excludes  academic laboratories with limited resources. In this work, we propose a new framework, dubbed ViCHA, that efficiently exploits the input data to boost the learning by: (a) a new hierarchical cross-modal alignment loss, (b) new self-supervised scheme based on masked image modeling, (c) leveraging image-level annotations, called \emph{Visual Concepts}, obtained with existing foundation models such as CLIP to boost the performance of the image encoder.  Although pretrained on four times less data, our ViCHA strategy outperforms other approaches on several downstream tasks such as Image-Text Retrieval, VQA, Visual Reasoning, Visual Entailment and Visual Grounding. The code will be made publicly available here: \href{https://github.com/mshukor/ViCHA}{https://github.com/mshukor/ViCHA}
\end{abstract}

\section{Introduction}
\label{sec:introduction}

Vision and Language Pretraining (VLP) \cite{du2022survey, chen2022vlpsurvey2} consists of training a vision and language model with simple pretraining tasks, like image-text alignment or masked language modelling, usually on large datasets. It is becoming the prominent paradigm to solve multimodal vision-language tasks (\emph{e.g.} VQA \cite{antol2015vqa}, NLVR$^2$ \cite{suhr-etal-2019-corpusnlvr}), outperforming other task-customised approaches \cite{ben2017mutan, ke2019reflective}. 
The representation learned by these models have shown to be useful also for unimodal tasks, paving the way for more general or foundational models \cite{singh2022flava, yuan2021florence, yu2022coca, radford2021learning}. 


Due to the abundance of image-text pairs data on the internet \cite{schuhmann2021laion, changpinyo2021conceptual, desai2021redcaps}, scaling these models has gained a lot of attention. The current most efficient approaches are those beyond 1B parameters \cite{yu2022coca, alayrac2022flamingo} and trained on hundred of millions to several billions examples \cite{radford2021learning, jia2021scaling, wang2022simvlm}. This race for bigger models/data, is most often done to the detriment of the quality of learning strategies, which becomes more and more difficult to control. We argue that a lot of improvements can be done by designing more efficient learning schemes. For example, while \citet{dosovitskiy2021anvit} train Vision Transformers with a huge pre-training dataset, \citet{touvron2021training} obtained state-of-the art results using the same architecture with bag of learning recipes, without pre-training.



\begin{figure}[t]
    \centering
    \includegraphics[width=\linewidth]{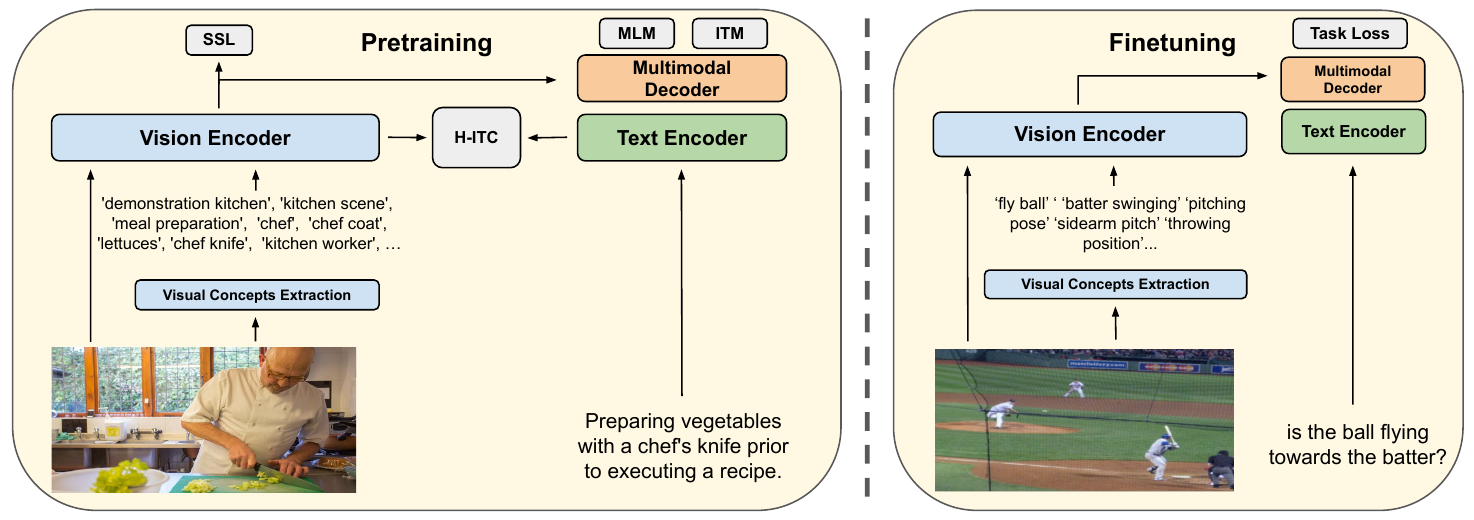}
    \vspace{-0.1cm}
    \caption{\small{Illustration of our ViCHA approach; during pretraining (left) and finetuning (right).}}
    \label{fig:main_overview}
    \vspace{-0.6cm}
\end{figure}

To this end, we propose a new VLP strategy with carefully designed training procedures and losses dedicated to efficiently align both modalities.
Specifically, we adopt an early fusion architecture, but instead of having only one alignment loss on top of the dual encoders, we also align both representations at several layers. This hierarchical alignment strategy is extended with a multimodal fusion decoder in order to capture more complex Vision-Language interaction. 
We also exploit more efficiently the input data with Masked Image Modeling, based on the recently proposed Masked Auto Encoder (MAE) \cite{mae}. 
%
%
Moreover, we leverage image-level annotations (\emph{i.e.}, objects and high level concepts describing the image) using existing foundation models (\emph{i.e}, CLIP \cite{radford2021learning}) due their generalization ability and low computational cost at inference time. Thus departing from classical use of such models (\emph{i.e.}, initialization and finetuning), and leveraging them without any retraining. We complete our scheme with a CLIP-based filtering technique.
An illustration of our approach, called \textbf{ViCHA}  for Efficient Vision-Language Pretraining with \textbf{Vi}sual \textbf{C}oncepts and \textbf{H}ierarchical \textbf{A}lignment, is presented in~Fig. \ref{fig:main_overview}. We will show the effectiveness of our approach on classical downstream tasks used for VLP evaluation while drastically limiting the size of the training datasets, contrary to other methods.



\section{Related Work}
\label{sec:related_work}
\noindent\paragraph{Vision and Language Pretraining (VLP):} 
VLP methods can be categorized into 3 main categories based on their model architectures. First, dual stream approaches have two separate encoders for images and texts (e.g. CLIP \cite{radford2021learning}, ALIGN \cite{jia2021scaling}) that are usually trained using a contrastive loss \cite{oord2018representation} on a global similarity between their output embeddings. These models are efficient during inference, but can not exploit finegrained cross modal interaction that is useful for multimodal tasks (\emph{e.g.}, VQA). 
 Several attempts based on self supervision \cite{mu2021slip, li2021supervision}, teacher-student distillation \cite{wang2021distilled, andonian2022robust} or finegrained interaction \cite{yao2021filip} have been proposed to alleviate the massive amount of data usually used for training (\emph{e.g.}, CLIP 400M). Second, unified approaches try to simplify the model and adopt one transformer that can process the two modalities \cite{wang2021vlmo, wang2021ufo}. These approaches are simpler, but still under perform other approaches with modality specific modules. Third, most of the work concentrated recently on hybrid approaches due to their success on multimodal tasks, where the models have separate encoders as well as a multimodal module that can exploit the cross-modal interaction more effectively. Early methods have relied on object detection models to extract image regions and tags \cite{lu2019vilbert, li2019visualbert, chen2020uniter, su2019vlbert, li2020unicoder, zhang2021vinvl}, however, the visual representation is limited by the performance of the pretrained object detector, which is expensive to scale on large annotated datasets, besides being slow at inference time. To remedy this, many methods have proposed to replace the object detector with vision transformers \cite{kim2021vilt, dou2021empiricalmeter, xue2021probingparsing} or CNNs \cite{huang2020pixel, wang2022simvlm}. These models are usually trained with image-text matching (ITM \cite{chen2020uniter}) or masked language modeling (MLM \cite{devlin2018bert}) losses, and recently with image text contrastive (ITC) to align the vision and language representation before feeding them to the fusion module \cite{li2021albef, yang2022visiontriplecont, duan2022multi}.
\noindent\paragraph{Multilevel Alignment:} Using multiscale representation extracted from different stages of the model have shown to be successful for vision tasks, in particular, some work in representation learning have proposed to do contrastive learning \cite{pissas2022multi} or mutual information maximization \cite{bachman2019learning} between features extracted at different scales, or between local and global features \cite{hjelm2018learning}.
In the context of VLP, \citet{li2022mvp} propose a multilevel semantic alignment by aligning local and global representations. \citet{gao2022pyramidclip} propose an intra and cross level alignment based on features extracted from cropped images and image regions on one side, and the caption and its summarization on the other side. \citet{li2021semvlp} use both single and dual stream encoders to align the two modalities at multiple levels.
Our approach use simpler strategy to align the features of both modalities at different transformer blocks. 
\noindent\paragraph{Object tags:} Several works have explicitly used high level concepts for vision-language tasks (\emph{e.g.} for image captioning and VQA \cite{wu2016value, you2016image}).  In the context of VLP, the concepts are usually extracted from an off-the-shelf object detectors \cite{ren2015faster} which are fed to the multimodal transformer, alongside region features and text tokens \cite{zhou2020unified}. Other than using object detectors, some approaches extract the concepts from the captions using Scene Graph Parser \cite{anderson2016spice}. These concepts are used in several ways; predicting these concepts \cite{tan-bansal-2019-lxmert, liu2021kd, yu2021ernie}, using them as a bridge between non aligned image and text data \cite{li2020unsupervised}, adopting a more efficient masking strategies \cite{cui2021rosita} or leveraging them for fine-grained alignment \cite{Wu_2019_CVPR_VSE}. In particular, OSCAR \cite{li2020oscar} uses the object tags to ease the alignment process by considering them as anchors for masked tokens prediction and contrastive learning.  The concepts extracted from pretrained object detection models are limited to the training classes, in addition, scene graph parser are noisy and there is no guaranty that all the concepts in the caption are extracted. Our approach leverages a more general model to capture a large set of diverse concepts and using them to enhance the visual representation.
\noindent\paragraph{Leveraging Foundational Models:} 
In recent years the focus have been shifted from task-customized models to more general or foundational models \cite{singh2022flava, yuan2021florence, radford2021learning}. The motivation is to have a holistic model that can be later used or adapted to many downstream tasks. 
Due to their generalization capabilities, models such as CLIP \cite{radford2021learning},  have been successfully leveraged in many tasks and domains, such as image editing \cite{couairon2022flexit}, generation \cite{ramesh2022hierarchicaldalle2}, segmentation \cite{wang2022cris}, captioning \cite{su2022language}, food retrieval \cite{shukor2022transformertfood}, explainability \cite{sammani2022nlx} and beyond. In the context of VLP, these models are used as initialization \cite{shen2022how} and recently, for dataset filtering \cite{schuhmann2021laion}. In this work, we leveraged CLIP for the extraction of VCs and data selection/filtering.

\section{ViCHA framework}
\label{sec:method}

\begin{figure}[t]
    \centering
    \includegraphics[width=\linewidth]{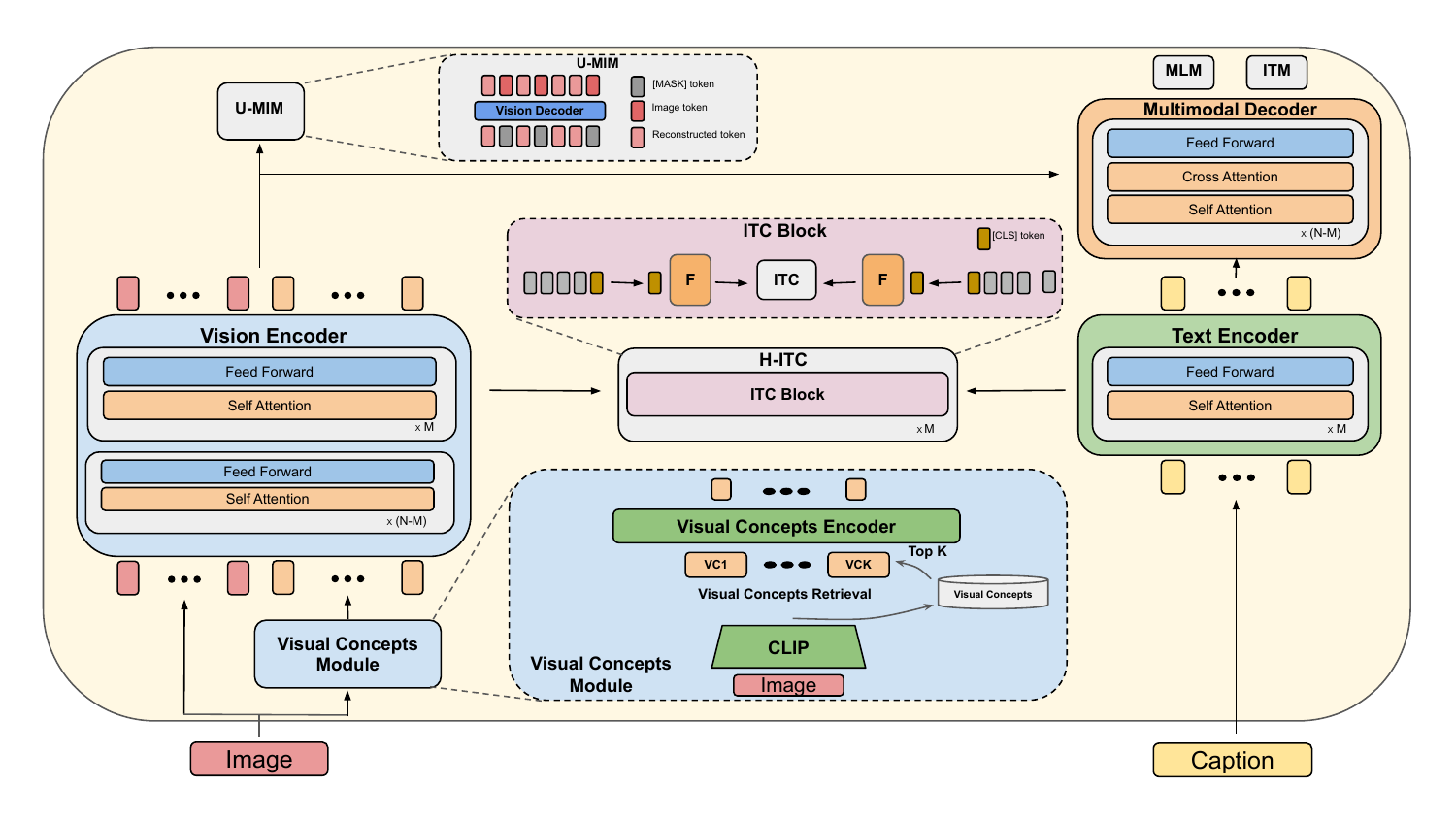}
    \caption{\small{ViCHA framework: our Vision-Language transformer model consists of the vision encoder (left) and the text encoder (right), both aligned through our Hierarchical Image-Text Contrastive block (H-ITC). Importantly, the input visual tokens are completed through the Visual Concepts Module providing extra visual semantic tokens. Finally, a multimodal decoder (top right) allows to learn complex multimodal relationship thanks to MLM and ITM objectives. We also introduce a unimodal Masked Image Modeling (U-MIM) on top of the vision encoder.
    }}
    \label{fig:main}
\end{figure}

\noindent
Illustrated in Figure \ref{fig:main}, our scheme presents Vision and Text encoders with three original components: a Visual Concepts Module to enrich the image encoder with relevant VCs, a new cross-modal interaction to align visual and langage feature representations at multiple levels, and a self-supervised component based on the recently proposed Masked Auto Encoder.
\paragraph{Notations:} Given a dataset of image-text pairs $\{I_i, T_i\}_i^K$,
 each image $I_i$ is associated with several Visual Concepts (VCs) $\{C_i^1, ...,C_i^P\}$ extracted using CLIP and exploited by our vision encoder $E_v$. The VCs are projected to the image patches embedding space using a Visual Concept Encoder $E_{vc}$. $E_v$ is a vision transformer that takes the image and the concepts as input and extracts the image tokens alongside a special class token $\{v_{cls}, v_1, ...,v_M\}$. Similarly, the text transformer encoder $E_t$ takes the text or image caption $T$ and extracts the text tokens $\{t_{cls}, t_1, ...,t_N\}$. The class tokens are aligned using a hierarchical contrastive loss before feeding the image and text tokens to a multimodal transformer decoder $E_{vl}$ that takes the text tokens as query and the image ones as keys and values. 

\subsection{Enhancing Visual Representation with Visual Concepts}

We propose to enhance the visual representation by explicitly injecting visual concepts. These concepts are extracted for each image, projected using a visual concepts encoder $E_{vc}$ and then concatenated to the patch tokens before feeding them to the vision transformer $E_v$. 

The motivation behind VCs is two-fold; VCs (a) might guide the vision encoder to focus on important objects/aspects of the image, (b) facilitate the alignment with the textual modality, first, because the visual tokens are already fused with the textual tokens of the VCs, second, it is easier to align the caption/text with the VCs tokens than the image tokens, especially at the beginning of the training.

Also we can see this technique from a visual prompt perspective. Text prompts are usually used with large language models to steer their output based on a given task/input \cite{liu2021pre, yao2022prompt, shin2020autoprompt, brown2020language}. We derive the analogy between these prompts and the prepend of VCs, as they may also guide the model to focus on different aspects of the image during training. 

Next, we detail how we extract, project and incorporate the VCs in our framework;

\paragraph{Visual Concepts Extraction:}
 Different from other approaches (\emph{e.g.}, OSCAR \cite{li2020oscar}) that extract these concepts (\emph{i.e.}, object tags) using pretrained object detectors, we use an off-the-shelf foundational model to extract more diverse, larger and semantically meaningful set of visual concepts. We show in section \ref{sec:results}, that this approach helps to capture high level and global concepts describing the scenes, which are hard to obtain with other approaches (\emph{e.g.} object detectors). The extraction of the visual concepts is done as follows;\vspace{-0.2cm}
\begin{enumerate}
    \item Concepts extraction: we extract the objects using Scene Graph Parsers \cite{schuster-etal-2015-generating_sgp} from all the captions of a given dataset. We use only simple text filters (transform to lower case and select objects repeated more than once). However, more complicated filters might improve the results even further.
    \item Concepts and image embeddings: We use a pretrained CLIP (ViT-B/16) model to obtain the embeddings of all the images and the extracted concepts.
    \item Concepts selection: for each image, we compute its cosine similarity with all the embedded concepts and select the top $k$ similar concepts. 
\end{enumerate}
\paragraph{Visual Concepts Encoder ($E_{vc}$):}
To encode Visual Concepts, we use a small text encoder $E_{vc}$ and then concatenate the output token embeddings with the image patch tokens before feeding $E_v$.

\paragraph{Visual Concepts Augmentation (VCA):} Here we propose a simple yet effective VCs augmentation technique. Having a set of concepts extracted for each image, instead of considering all the concepts at once, we sample randomly a fraction of these concepts (\emph{i.e.}, $p_{vc}$\%) at each iteration step. The benefit of VCA is three-fold; (1) it may prevent the model from overfitting on specific concepts and potentially disregard the image or other concepts during training, (2) the model sees different combinations of concepts, which helps to have more diversity. (3) the model is exposed to more information during finetuning or test, as it will see all the concepts at once.


\subsection{Training scheme}
We pretrain the model using several objectives; Hierarchical Image-Text Contrastive Learning (H-ITC), ITM, MLM and Masked Image Modeling (MIM).

\paragraph{Hierarchical Image-Text Contrastive Learning (H-ITC)}
Here we propose to exploit the hierarhical representations capture by the vision and language encoders. CNN based vision encoders learn hierarchical representations, starting from local features to more abstract ones at later stages \cite{zeiler2014visualizing}. Vision transformers have also been shown to display some level of hierarchy in learned representations \cite{dosovitskiy2021anvit, raghu2021vision}.


On the text side, recent works show that the attention heads specialize in particular part-of-speech tags, which differs across heads and layers \cite{vig2019analyzing} and reflect different aspects of the syntax \cite{clark2019does}, while the concepts learned by BERT differ significantly across layers and evolve into representing a linguistic hierarchy \cite{dalvi2022discovering}. 

Motivated by this, and different from other work \cite{li2021albef, yang2022visiontriplecont, duan2022multi} that align the two modalities only at the last layer, we propose to align them at different layers of the vision and text transformers. We argue that doing the alignment at early stages facilitates the process at the subsequent layers, while allowing to align the representation at different semantic levels. 

There is an asymmetry between the visual and textual information, as the image contains more diverse and detailed information while the caption usually contains  more abstract one, thus we align the textual features only with the visual ones from the last layers of $E_v$.

Specifically, at layer $i$, we compute the image-to-text and text-to-image similarities
and then apply a Softmax with a learned temperature parameter $\tau$:
\begin{equation}
    p_{i2t}^i = \frac{exp(s(I, T)^i/\tau)}{\sum_{m=1}^Q exp(s(I, T_m)^i/\tau)}, \quad p_{t2i}^i = \frac{exp(s(T, I)^i/\tau)}{\sum_{m=1}^Q exp(s(T, I_m)^i/\tau)},
\end{equation}
$T_m$ and $I_m$ are the negative text and image examples. $s(\cdot,\cdot)^i$ is the cosine similarity  and is obtained after linearly projecting and normalizing the class tokens at layer $i$ into a shared latent space:
\begin{equation}
    s(I, T)^i = g_{vi}(v_{cls}^i)^T g'_{ti}(t_{cls}^i), \quad s(T, I)^i = g_{ti}(t_{cls}^i)^T g'_{vi}(v_{cls}^i),
\end{equation}
where $g_{ti}()$ and $g_{vi}()$ are the linear projection layers at layer $i$. We maintain two queues of size $Q$ that store the normalized features from the momentum models $g'_{li}()$ and $g'_{vi}()$. We didn't see a significant improvement when using queues for the other layers, thus we use the queues only for the last layer for simplicity (on select in-batch negative examples for other layers). We then compute the cross entropy loss  between $\bm{p}$ and the one-hot ground truth and sum across all layers.

\paragraph{Image-Text Matching (ITM)} ITM loss is applied on top of the multimodal trasformer decoder for more finegrained fusion. It is a binary cross entropy loss, where the model tries to classify if the image-text pairs are positive or negative. The image and its corresponding caption are considered as positive pairs while the other examples in the batch are considered as negative. 
We sample a hard text and hard image from the batch based on the global cosine similarity on top of the dual encoders \cite{li2021albef}. 

\paragraph{Masked Language Modeling (MLM)} MLM consists of predicting a masked token given other contextual tokens. In our work, the model has also access to the visual tokens, which helps to learn a cross modal representation. Following BERT \cite{devlin2018bert}, we mask 15\% of the tokens and replace them with [MASK] token, random token or we keep them unchanged with probabilities of 80\%, 10\% and 10\% respectively. The task is a classification loss where the model should predict the token id from a list of vocabulary. 

\begin{table*}[!t]
    \small
	\centering	
 \setlength\tabcolsep{4pt}
	\resizebox{0.7\textwidth}{!}{%
	\begin{tabular}	{l  c | c  c  c  c  c  c}
		\toprule
	 \multirow{2}{*}{Method} & \# Pre-train &  \multicolumn{2}{c}{VQA} & \multicolumn{2}{c}{NLVR$^2$} & \multicolumn{2}{c}{SNLI-VE}  \\
	 & ~~Images & test-dev & test-std & dev & test-P & val & test \\
	 
	& & & & & & & \\
	\midrule
	ViLBERT~\cite{lu2019vilbert} & 3M &  70.55 & 70.92 & - & - & - & - \\ 
	l2-in-1~\cite{lu202012l2in1} & 3M &  73.15 & - & - & 78.87 & - & 76.95  \\
	ERNIE-ViL~\cite{yu2021ernie} & 3.8M &  73.18 & 73.36 & - & - & - & - \\
	ImageBERT~\cite{qi2020imagebert} & 6M & - & - & - & - & - & -  \\

	Unicoder-VL~\cite{li2020unicoder} & 3.8M & - & - & - & - & - & -  \\
	UNITER \cite{chen2020uniter} & 4M & 72.70 & 72.91 & 77.18 & 77.85 & 78.59 & 78.28  \\
	OSCAR \cite{li2020oscar} & 4M  &  73.16 & 73.44 & 78.07 & 78.36 & - & - \\
	VILLA \cite{gan2020largevilla} & 4M & 73.59 & 73.67 & 78.39 & 79.30 & 79.47 & 79.03 \\
	UNIMO\_B \cite{li2020unimo} & 4M+1.7M & 73.79 & 74.02 & - & - & 80.00 & 79.10  \\ 
	\midrule
	ViLT \cite{kim2021vilt} & 4M & 70.94 & - & 75.24 & 76.21& - & -  \\
	ALBEF \cite{li2021albef} & 4M & 74.54 & 74.70 & 80.24 & 80.50 & 80.14 & 80.30  \\

	\midrule
	ALBEF$^*$ & 1.1M & 72.51 & 72.69 & 75.72 & 76.31 & 78.08 & 78.02 \\
	ViCHA  & 1.1M & \textbf{73.55} & \textbf{73.52} & \textbf{77.27} & \textbf{77.08} & \textbf{78.96} & \textbf{78.22}  \\
	\midrule
	ViCHA$^\dagger$  & 800K & 73.23   &  -  &  78.14  &  77.00  &  79.02 &  78.65    \\
		\bottomrule
	\end{tabular}
	}
	\vspace{2ex}
	\caption
	{
	\footnotesize Comparison with SOTA; we report the accuracy on VQA, NLVR$^2$ and VE. ViCHA outperforms other approaches trained on much more data ($\sim$ 4M images) and SOTA trained on the same setup (ALBEF$^*$). Our model trained on the filtered dataset (ViCHA$^\dagger$) is very competitive while using much less data ($\sim$ 800K images). Results on 4M images are in the supp.
	}
    \vspace{-0.5cm}
	\label{tab:vqa}

\end{table*}

\paragraph{Masked Image Modeling (MIM):} Improving the visual representation have shown to affect significantly the performance on vision-language tasks \cite{zhang2021vinvl}. Many work in self supervised learning (SSL) have been proposed \cite{he2020momentummoco, caron2021emergingdino}, in particular, the methods based on masked image reconstruction have achieved SOTA results \cite{bao2022beit, mae}. However, it is not clear how much MIM is useful for VLP, as recent work show that MIM based on approaches such as BEiT, masked region regression or classification degrades the performance \cite{dou2021empiricalmeter, kim2021vilt, hendricks-etal-2021-decoupling}. Motivated by these findings, here we propose two approaches to investigate whether MIM could help VLP. The two methods are based on the recent Masked Auto Encoder (MAE) approach \cite{mae}, where we randomly mask the image (\emph{e.g.}, 75\%) and pass only the unmasked tokens to $E_v$, however, they are different in how they reconstruct the image:

\paragraph{Unimodal MIM (U-MIM):} Here we propose to improve the visual representation by reconstructing the masked tokens given only the unmasked image ones (without access to VCs). Specifically, we add a small decoder $D_v$ (\emph{e.g.}, 2-layer transformer) that takes the output tokens, concatenated to the masked ones ($\bm{\hat{v}}$) and reconstruct the input image as $\hat{I}$. The unimodal masked image modeling (U-MIM) loss can be written as:
\begin{equation}
    \mathcal{L}_{U-MIM} = MSE(I, D_v(\bm{\hat{v}})).
\end{equation}

\noindent\textbf{Multimodal MIM (M-MIM):} Here the decoder has access also to the textual tokens, which helps to provide more informative context to reconstruct the masked tokens. The decoder is the same as our $E_{vl}$ (shared weights) but is provided with the visual tokens ($\bm{\hat{v}}$) as queries and the textual ones ($\bm{t}$) as keys and values (hence, the queries, keys and values are different when computing MLM and ITM). The loss can be written as:
\begin{equation}
    \mathcal{L}_{M-MIM} = MSE(I, F(E_{vl}(\bm{\hat{v}}, \bm{t}))),
\end{equation}
where $F$ is a linear projection.
The total loss can be written as ($\lambda_{H-ITC}=0.1$ and $\lambda_{MIM}=1$):
\begin{equation}
    \mathcal{L} = \mathcal{L}_{ITM} + \mathcal{L}_{MLM} + \lambda_{H-ITC}\mathcal{L}_{H-ITC} + \lambda_{MIM}\mathcal{L}_{MIM}.
\end{equation}

\begin{table*}[!t]
    \small
	\centering	
 \setlength\tabcolsep{4pt}
	\resizebox{0.9\textwidth}{!}{%
	\begin{tabular}	{l  c | c  c  c  c  c  c | c  c  c  c  c  c }
		\toprule
	 \multirow{2}{*}{Method} & \# Pre-train &  \multicolumn{6}{c|}{Flickr30K (1K test set)} & \multicolumn{6}{c}{MSCOCO (5K test set)} \\
	 & ~~Images & \multicolumn{3}{c}{TR}& \multicolumn{3}{c|}{IR} &  \multicolumn{3}{c}{TR}& \multicolumn{3}{c}{IR}\\
	 
	& & R@1 &R@5&R@10& R@1 &R@5&R@10& R@1 &R@5&R@10& R@1 &R@5&R@10\\
	\midrule
	ViLBERT~\cite{lu2019vilbert} & 3M &  - & - & - & 58.2 & 84.9 & 91.5 & - &- &- &- &- &- \\ 
	l2-in-1~\cite{lu202012l2in1} & 3M  & - & - & - & 67.90 & - & - & - &- &- &68.00 &- &- \\
	ERNIE-ViL~\cite{yu2021ernie} & 3.8M &  86.7 & 97.80 & 99.00 & 74.44 & 92.72 & 95.54 & - &- &- & - &- &- \\
	ImageBERT~\cite{qi2020imagebert} & 6M &  87.0 & 97.6 & 99.2 & 73.1 & 92.6 & 96.0 & 66.4 & 89.8 & 94.4 & 50.5 & 78.7 & 87.1 \\

	Unicoder-VL~\cite{li2020unicoder} & 3.8M & - & - & - & - & - & - & 62.3 & 87.1 & 92.8 & 46.7 &76.0 &85.3 \\
	UNITER \cite{chen2020uniter} & 4M  & 87.3 & 98.0 &99.2 &75.6 &94.1 &96.8 &65.7 &88.6 &93.8 &52.9 &79.9 &88.0 \\
	OSCAR \cite{li2020oscar} & 4M   & - & - & - & - & - & - & 70.0 & 91.1 & 95.5 &  54.0 & 80.8 & 88.5\\
	VILLA \cite{gan2020largevilla} & 4M  & 87.9 & 97.5 & 98.8 & 76.3 & 94.2 & 96.8 & - & - & - & - & - & - \\
	UNIMO\_B \cite{li2020unimo} & 4M+1.7M  & 89.7 & 98.4 & 99.1 & 74.7 & 93.4 & 96.1 & - & - & - & - & - & -  \\ 
	\midrule
	ViLT \cite{kim2021vilt} & 4M  & 83.5 & 96.7 & 98.6 & 64.4 & 88.7 & 93.8 & 61.5 &86.3 &92.7 &42.7 &72.9 &83.1 \\
	ALBEF \cite{li2021albef} & 4M  & 94.3 & 99.4 & 99.8 & 82.8 & 96.7 & 98.4 & 73.1 &91.4 &96.0 &56.8 &81.5 &89.2 \\

	\midrule
	ALBEF$^*$ & 1.1M  & 88.1 & 97.8 & 99.0 & 73.6 & 92.0 & 95.6 & 71.2 & 91.3 & 95.6 & 54.2 & 80.7 & 88.6 \\
	ViCHA  & 1.1M &  \textbf{91.7}  & \textbf{98.7}  & \textbf{99.5}  & \textbf{77.2} & \textbf{94.2} & \textbf{96.8} & \textbf{73.6} & \textbf{92.4} & \textbf{96.2} & \textbf{56.8} & \textbf{82.2}  & \textbf{89.5} \\
	\midrule
	ViCHA$^\dagger$  & 800K &  90.0  & 98.4  & 99.8  & 77.4 & 94.3 & 96.7 & 73.3  & 92.1  & 96.2  & 55.8  &  81.8  & 89.1  \\
		\bottomrule
	\end{tabular}
	}
	\vspace{2ex}
	\caption
	{
	\footnotesize Comparison with SOTA; we report R@K for finetuning on Image-Text Retrieval. ViCHA outperforms other approaches trained on much more data ($\sim$ 4M images) and SOTA trained on the same setup (ALBEF$^*$). Our model trained on the filtered dataset (ViCHA$^\dagger$) is very competitive while using much less data ($\sim$ 800K images). Results on 4M images are in the supp.
	}
    \vspace{-0.5cm}
	\label{tab:ret}

\end{table*}

\section{Experiments}
\label{sec:results}
We follow other works \cite{li2021albef} and evaluate the model on four downstream tasks: Image-Text Retrieval, VQA, NLVR$^2$, Visual Entailment and Visual Grounding. More details about downstream tasks, implementation details, comparability and dataset filtering can be found in the Appendix. 

\paragraph{Our setup:} We favor low data and compute regimes. Specifically, we pretrain only on 3 public datasets; COCO \cite{lin2014microsoftcoco}, Visual Genome \cite{krishna2017visualgenome} and SBU \cite{sbu}, which account to $\sim 1.1M$ images in total, thus, 4 times lower than other approaches (\emph{e.g.} 4M images \cite{li2021albef, chen2020uniter, dou2021empiricalmeter}). In addition, we train for 10 epochs (in contrast to 30 epochs \cite{li2021albef, yang2022visiontriplecont, duan2022multi}) with relatively small batch size of 128 (32 per GPU) using 4 GPUs.

\paragraph{Filtered dataset:} We go further and apply a CLIP-based filtering technique to reduce the number of images to $\sim 800k$, and train our approach on this dataset (ViCHA$\dagger$). This dataset consists of; COCO, 50\% of VG captions and 70 \% of SBU.

\paragraph{Implementation details:}\label{sec:implem_details} We follow the implementation of ALBEF \cite{li2021albef}, including the same architecture, with our setup. The visual encoder is ViT-B/16 \cite{dosovitskiy2021anvit, touvron2021training}, the text encoder is the first 6 layers of BERT-base \cite{devlin2018bert} and the multimodal encoder is the last 6 layers of BERT-base. $E_{vc}$ is the first 2 layers of BERT-base. We extract 15 concepts for each image and we set $p_{vc}$=30\% for VCA. For H-ITC loss, we use the last 6 layers of $E_v$ and the all 6 layers of $E_l$. For U-MIM, we use 2-layer transformer encoder.
 
To have a fair comparison, we compare with ALBEF trained on the same setup (called ALBEF$^*$). Even though it is common to compare among all approaches in the literature, we think that it is hard to assess the methods as they follow different training setups.

\paragraph{Comparison with SOTA on standard tasks:} Table \ref{tab:vqa} shows a comparison with other approaches, on \emph{VQA, NLVR$^2$, VE} following the finetuning setup. We outperform significantly ALBEF$^*$ (+1.04\% VQA, +1.55\% NLVR$^2$ and +0.88\% VE) and other approaches trained on more data (\emph{i.e.}, 4M images), such as ViLT (+2.61\% VQA, +2.03\% NLVR$^2$), UNITER (+0.85\% VQA, and +0.37\% VE) and OSCAR (+0.39\% VQA). For \emph{Image-Text Retrieval}, The model is evaluated on COCO and Flickr30K (F30K) following 2 setups; finetuning and zero-shot. Table \ref{tab:ret} shows the finetuning results. Compared to ALBEF$^*$, our approach achieves significant improvements, especially on R@1 with absolute improvement of 3.74\% IR and 3.6\% TR on Flickr30K, and 2.59\% IR and 2.42\% TR on the more challenging COCO. Interestingly, we outperform other approaches trained on more data such as ViLT, UNITER and UNIMO (+2.58\% R@1 IR F30K). Compared to the SOTA trained with 4M images, we outperform ALBEF \cite{li2021albef} on COCO (+1.72\% TR RSUM and +0.88\% IR RSUM). On zero-shot F30K, we follow other approaches \cite{li2021albef, yang2022visiontriplecont, duan2022multi} and use the model finetuned on COCO. While for zero-shot COCO, we directly use the model after pretraining (without VCA). Note that, by zero-shot, we mean the model is not explicitly trained on the target dataset after pretraining. Table \ref{tab:retrieval_zs} shows that our model significantly outperforms ALBEF$^*$ (+7.88\% F30K and +8.76\% COCO R@1 IR) and other approaches trained on more data such as UNITER, ViLT, CLIP on F30K IR and COCO, as well as ALIGN on COCO.  

\begin{table*}[!t]
    \small
	\centering	
 \setlength\tabcolsep{4pt}
	\resizebox{0.8\textwidth}{!}{%
	\begin{tabular}	{l  c |  c  c  c  c  c  c | c  c  c  c  c  c }
		\toprule	 	
	 \multirow{2}{*}{Method} & \# Pre-train & \multicolumn{6}{c|}{Flickr30K (1K test set)} & \multicolumn{6}{c}{MSCOCO (5K test set)} \\
	 & ~~Images &  \multicolumn{3}{c}{TR}& \multicolumn{3}{c|}{IR} &  \multicolumn{3}{c}{TR}& \multicolumn{3}{c}{IR}\\
	 \midrule
	& & R@1 &R@5&R@10& R@1 &R@5&R@10& R@1 &R@5&R@10& R@1 &R@5&R@10\\
	ImageBERT~\cite{qi2020imagebert} & 6M & 70.7 & 90.2 & 94.0 & 54.3 & 79.6 & 87.5 & 44.0 & 71.2 & 80.4 & 32.3 & 59.0 & 70.2  \\
	UNITER \cite{chen2020uniter} & 4M & 80.7& 95.7& 98.0& 66.2& 88.4& 92.9 &- &- &- &- &- &- \\
	ViLT \cite{kim2021vilt} & 4M & 73.2 & 93.6 & 96.5 & 55.0 & 82.5 & 89.8 & 56.5 & 82.6 & 89.6 & 40.4 & 70.0 & 81.1 \\
	ALBEF \cite{li2021albef} & 4M & 90.5 & 98.8 &
99.7 & 76.8 & 93.7 & 96.7 & 68.7 & 89.5 & 94.7 & 50.1 & 76.4 & 84.5 \\
    \midrule
    CLIP \cite{radford2021learning} & 400M & 88.0 & 98.7 & 99.4 & 68.7 & 90.6 & 95.2 & 58.4 & 81.5 & 88.1 & 37.8 & 62.4 & 72.2 \\
	ALIGN \cite{jia2021scaling} & 1.2B& 88.6 & 98.7 & 99.7 & 75.7 & 93.8 & 96.8 & 58.6 & 83.0 & 89.7 & 45.6 & 69.8 & 78.6 \\
	\midrule
	ALBEF$^*$  & 1.1M & 81.3  & 94.9  & 98.2  & 64.7  & 87.2 & 92.0  & 59.5 & 84.4 & 91.2 & 38.3 & 68.0 & 79.1 \\

	ViCHA  & 1.1M& \textbf{86.0}  & \textbf{97.1}  & \textbf{99.0} & \textbf{72.6}  & \textbf{91.1}  & \textbf{95.0}  &\textbf{63.1} & \textbf{86.9} & \textbf{92.7} & \textbf{47.1} & \textbf{73.8} & \textbf{82.8} \\
	\midrule
	ViCHA$^\dagger$  & 800K & 85.0   &  96.8   & 98.7    & 71.6 & 91.0    & 94.9 & - & -  &  -  &  -  &   - & -   \\
		\bottomrule
	\end{tabular}
	}
	\vspace{2ex}
	\caption
	{
	\footnotesize	
		Zero-shot Comparison with SOTA on Flickr30K (after fine-tuning ViCHA on COCO) 
		and COCO (pretrained model only).
		}
	\label{tab:retrieval_zs}

\end{table*}

\paragraph{Ablation Study:} Table \ref{tab:ablation_main} shows the contribution of the different components of our ViCHA strategy. Using VCs seems to give significant improvements over the baseline (+9.6\% RSUM and +0.5\% Acc. VQA). Then we add our H-ITC loss, which also brings additional points (+2.2\% RSUM and +0.2\% Acc. VQA), showing the importance of a stronger pre-alignment loss. We favor the MAE based, unimodal MIM loss which improves the results by 2.4\% RSUM. We choose U-MIM over M-MIM due to its superior performance, however, both objectives gives better result than the baseline; +7.8\% and +3.2\% RSUM for U-MIM and M-MIM respectively (detailed results can be found in the appendix). This reveals the importance of MIM as well as using SSL objectives for VLP. We finally add VCA, that significantly help the VQA task (+0.9\% Acc.). Overall, we show that all contributions are not antagonistic and can be combined effectively. More ablation can be found in the appendix.

\begin{table*}[!t]
    \small
	\centering	
 \setlength\tabcolsep{4pt}
	\resizebox{0.75\textwidth}{!}{%
	\begin{tabular}	{c  c  c  c |  c  c  c  c  c  c  c  | c }
		\toprule	 	
	 \multirow{3}{*}{VCs} & \multirow{3}{*}{H-ITC} & \multirow{3}{*}{U-MIM} & \multirow{3}{*}{VCA} & \multicolumn{7}{c|}{Flickr30K (1K test set)} & VQA \\
	 & & &  &  & \multicolumn{3}{c}{TR}& \multicolumn{3}{c}{IR} & test-dev\\
	& & & & RSUM & R@1 &R@5&R@10& R@1 &R@5&R@10 & Acc.\\
	\midrule
    \multicolumn{4}{c|}{Baseline (ALBEF$^*$)} &  535.8 & 85.8 & 97.4 & 98.5 & 70.2 & 89.9 & 94.0 & 71.1 \\
    \midrule
	\ding{51} & &  &  & 545.4&  87.6 & 97.6 & 99.3 & 74.0 & 91.6 & 95.3 & 71.6 \\
	 & \ding{51} &  &  & 545.1 & 87.3 & 98.1 & 99.3 & 73.2 & 92.0 & 95.2 & 72.0  \\
	 &  &  \ding{51} & & 543.6 & 88.6 & 97.6 & 99.2 & 72.3 & 91.1 & 94.8 & 71.8\\ \midrule
	 \ding{51} & \ding{51} &  &  & 547.6 & 88.3 & 97.6 & 99.3 & 73.7 & 92.7 & 96.0  & 71.8 \\ 
	 \ding{51} & \ding{51} & \ding{51}  &  & 550.0& 89.8 & 97.6 & 99.3 & 74.9 & 92.8 & 95.6 & 71.7 \\
    \ding{51} & \ding{51} & \ding{51} &  \ding{51} &  550.7 & 88.2  & 97.8  & 99.4  & 75.7  & 93.4  & 96.2 & 72.6 \\	 
    \bottomrule

	\end{tabular}
	}
	\vspace{2ex}
	\caption
	{
	\small	
		Ablation study: The models are finetuned on Flickr30K and VQA v2 (as in \cite{dou2021empiricalmeter, yang2022visiontriplecont}).
	}
    \vspace{-0.5cm}  
	\label{tab:ablation_main}

\end{table*}

\paragraph{Data and Compute Efficiency:} The way the VCs are extracted (using CLIP) is not central to our work, as for the gain coming from 400M pairs of CLIP. This is supported in our experiments in the supplementary material, where we replace CLIP by an object detector. In addition, Table \ref{tab:ablation_main} shows that other components contribute significantly, where the gain coming from H-ITC and U-MIM individually, is comparable to VCs. The number of params of ViCHA during inference is 248.5 M compared to 209.9 M for ALBEF and $\sim$ 265 M for METER \cite{dou2021empiricalmeter}. The training time of ViCHA (1M) is $\sim$ 70 hours, for ViCHA (800K) the training time is much lower (39h). This is with the paper setup (4 GPUs A100). Compared to other SOTA, using 8GPUs A100; ALBEF \cite{li2021albef} takes 2-3 days and METER 8 days \cite{dou2021empiricalmeter}.

\paragraph{Additional Scores on Visual Grounding:} Table \ref{tab:grounding} shows a comparison between Weakly Supervised Visual Grounding (WSVG) approaches on the RefCOCO+ dataset. We outperform ALBEF$^*$ (+0.67 \% val, +1.06 \% TestA and + 1.19 \% TestB) as well other approaches such as DTMR \citet{sun2021discriminative} and KPRN \citet{liu2019knowledge}. Compared to ALBEF trained on 4M images we obtain better performance on the TestA set (+ 0.68 \%). In addition, we show some qualitative results using Grad-CAM by back propagating the ITM loss until the 3rd layer of the multimodal encoder, the results can be found in the appendix. 

\paragraph{Training on 4M images:} In this section, we train ViCHA on 4M images and compare it with other SoTA. We follow the same implementation details described in the paper except the following (to have a fair comparison with other work \cite{li2021albef}); we add CC3M \cite{Sharma2018ConceptualCA}, pretrain for 30 epochs using 8 GPUs, during Finetuning we increase the image resolution to 384 for VQA and NLVR$^2$, we use additional Visual Genome questions for VQA. Figure \ref{tab:all_4m}, shows that ViCHA outperforms state of the art models on most of the tasks (\emph{i.e.}, VQA, NLVR-dev, F30K R@1 TR, MSCOCO R@1 TR) and competitive on other tasks. This validate the effectiveness of the proposed approach, however, the improvement is larger on low data regime (1M images) as shown in Tables \ref{tab:ret} and \ref{tab:vqa}.

\begin{table*}[!h]
    \small
	\centering	
 \setlength\tabcolsep{4pt}
	\resizebox{1\textwidth}{!}{%
	\begin{tabular}	{l  c | c  c  c  c  c  c| c  c  c  c  c  c | c  c  c  c  c  c }
		\toprule
	 \multirow{2}{*}{Method} & \# Pre-train &  \multicolumn{2}{c}{VQA} & \multicolumn{2}{c}{NLVR$^2$} & \multicolumn{2}{c}{SNLI-VE} & \multicolumn{6}{c|}{Flickr30K (1K test set)} & \multicolumn{6}{c}{MSCOCO (5K test set)} \\
	 & ~~Images & test-dev & test-std & dev & test-P & val & test & \multicolumn{3}{c}{TR}& \multicolumn{3}{c|}{IR} &  \multicolumn{3}{c}{TR}& \multicolumn{3}{c}{IR}\\
	 
	& & & & & & & &R@1 &R@5&R@10& R@1 &R@5&R@10& R@1 &R@5&R@10& R@1 &R@5&R@10\\
	\midrule
	ViLBERT~\cite{lu2019vilbert} & 3M &  70.55 & 70.92 & - & - & - & - & - & - & - & 58.2 & 84.9 & 91.5 & - &- &- &- &- &- \\ 
	l2-in-1~\cite{lu202012l2in1} & 3M &  73.15 & - & - & 78.87 & - & 76.95 & - & - & - & 67.90 & - & - & - &- &- &68.00 &- &- \\
	ERNIE-ViL~\cite{yu2021ernie} & 3.8M &  73.18 & 73.36 & - & - & - & - & 86.7 & 97.80 & 99.00 & 74.44 & 92.72 & 95.54 & - &- &- & - &- &- \\
	ImageBERT~\cite{qi2020imagebert} & 6M & - & - & - & - & - & - & 87.0 & 97.6 & 99.2 & 73.1 & 92.6 & 96.0 & 66.4 & 89.8 & 94.4 & 50.5 & 78.7 & 87.1 \\

	Unicoder-VL~\cite{li2020unicoder} & 3.8M & - & - & - & - & - & - & - & - & - & - & - & - & 62.3 & 87.1 & 92.8 & 46.7 &76.0 &85.3 \\
	UNITER \cite{chen2020uniter} & 4M & 72.70 & 72.91 & 77.18 & 77.85 & 78.59 & 78.28 & 87.3 & 98.0 &99.2 &75.6 &94.1 &96.8 &65.7 &88.6 &93.8 &52.9 &79.9 &88.0 \\
	OSCAR \cite{li2020oscar} & 4M  &  73.16 & 73.44 & 78.07 & 78.36 & - & - & - & - & - & - & - & - & 70.0 & 91.1 & 95.5 &  54.0 & 80.8 & 88.5\\
	VILLA \cite{gan2020largevilla} & 4M & 73.59 & 73.67 & 78.39 & 79.30 & 79.47 & 79.03 & 87.9 & 97.5 & 98.8 & 76.3 & 94.2 & 96.8 & - & - & - & - & - & - \\
	UNIMO$_B$ \cite{li2020unimo} & 4M+1.7M & 73.79 & 74.02 & - & - & 80.00 & 79.10 & 89.70 & 98.40 & 99.10 & 74.66 & 93.40 & 96.08 & - & - & - & - & - & -  \\ 
	\midrule
	ViLT \cite{kim2021vilt} & 4M & 70.94 & - & 75.24 & 76.21& - & - & 83.5 & 96.7 & 98.6 & 64.4 & 88.7 & 93.8 & 61.5 &86.3 &92.7 &42.7 &72.9 &83.1 \\
	ALBEF \cite{li2021albef} & 4M & 74.54 & 74.70 & 80.24 & 80.50 & 80.14 & 80.30 & 94.3 & 99.4 & 99.8 & 82.8 & \textbf{96.7} & 98.4 & 73.1 &91.4 &96.0 &56.8 &81.5 &89.2 \\

 	COTS \cite{lu2022cots} & 4M & - & - & - & - & - & - & 88.2 & 98.5 & 99.7 & 75.2 & 93.6 & 96.5 & 66.9 & 88.8 & 94.0 & 50.5 & 77.6 & 86.1 \\
  
 	CODIS \cite{duan2022multi} & 4M & 74.86 & 74.97 & 80.50 & 80.84 & 80.47 & \textbf{80.40} & 95.1 & 99.4 & 99.9 & 83.3 & 96.1 & 97.8 & 75.3 &92.6 &96.6 &58.7 &82.8 &89.7 \\
  
	TCL \cite{yang2022visiontriplecont} & 4M & 74.90 & 74.92 & 80.54 & 81.33 & \textbf{80.51} & 80.29 & 94.9 & \textbf{99.5} & 99.8 & \textbf{84.0} & \textbf{96.7} & \textbf{98.5} & 75.6 &92.8 &96.7 & \textbf{59.0} & \textbf{83.2} &89.9 \\

	MixGen \cite{hao2022mixgen} & 3M & 74.51 & 74.79 & 80.23 & \textbf{80.94} & 80.05 & 80.05 & 94.8 & 99.4 & \textbf{100} & 82.4 & 96.3 & 98.0 & 74.2 &92.8 &96.4 &57.3 &82.1 &89.0 \\
 
	\midrule
	ViCHA   & 4M & \textbf{74.99}   &  \textbf{75.07}  &  \textbf{80.84}  &  80.44  &  79.92 &  79.35  & \textbf{95.2}  & 99.1  & 99.8  & 82.56 & 95.56 & 97.88 & \textbf{77.2}  & \textbf{93.2}  & \textbf{96.7}  & 58.8  &  \textbf{83.2}  & \textbf{90.2}  \\
		\bottomrule
	\end{tabular}
	}
	\vspace{2ex}
	\caption
	{
	\footnotesize Comparison with SOTA; we report the accuracy on VQA, NLVR$^2$, VE and R@K for Image-Text Retrieval. ViCHA ouperforms SoTA on most of the tasks.
	}
	\label{tab:all_4m}

\end{table*}

\paragraph{Data filtering:} Interestingly, the model trained on the filtered data (ViCHA$^\dagger$) gives comparable results on image-text retrieval (Table \ref{tab:ret} and \ref{tab:retrieval_zs}), sometimes better than, training on all data (\emph{e.g.}, NLVR$^2$ and SNLI-VE, Table \ref{tab:vqa}), while being always better than ALBEF$^*$.

\begin{figure}[h]
\captionsetup[subfigure]{labelformat=empty}
    \centering
    \subfloat[\tiny{\textbf{Dataset}: COCO \\ \textbf{Caption}: "Prepping vegetables with a chef's knife prior to executing a recipe."\\
    \textbf{VCs}: 'demonstration kitchen', 'cooking demonstration', 'kitchen restaurant scene', 'galley kitchen', 'kitchen scene', 'type kitchen', 'meal preparation', 'food preparation', 'chef', 'chef coat', 'lettuces', 'chef knife', 'efficiency kitchen', 'plan kitchen', 'kitchen worker'}]{
        \includegraphics[width=0.25\textwidth, height=0.2\textwidth]{images/vcs/coco_3000.jpg}
    }\hspace{0.6cm}
    \subfloat[\tiny{\textbf{Dataset}: SBU \\ \textbf{Caption}: ' Mom opening up a bottle of cider mix from Catherine.  Stupid tree was in my way.'\\
    \textbf{VCs}: 'santa juana', 'mama santa', 'bev pelletier', 'pam tree', 'amazon', 'sue veeder', 'christmaspresents', 'christmas woman', 'christmas basket', 'anne peterson', 'santa gifts', 'holiday cheer', 'pamela hinshaw', 'cynthia rybakoff', 'christmas'}]{
        \includegraphics[width=0.25\textwidth, height=0.2\textwidth]{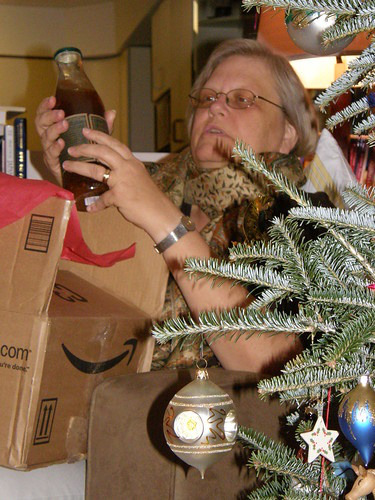}
    }\hspace{0.6cm}
    \subfloat[\tiny{\textbf{Dataset}: VG \\ \textbf{Caption}: 'the hat is pink'\\
    \textbf{VCs}:'birthday celebrate', 'birthday celebration', 'birthday party', 'cake cutting', 'celebration cake', 'cake portion', 'birthay cake', 'bithday cake', 'child cake', 'birth day', 'celebration', 'cutting cake', 'birthday', 'congratulations cake', 'birthday cake'}]{
        \includegraphics[width=0.25\textwidth, height=0.2\textwidth]{images/vcs/vg_0.jpg}
    }\\
    \caption{Illustration of VCs: VCs capture global information, actions and other aspects of the scene that help to give a rich context.}
    \label{fig:vcs_general}
\end{figure}

\paragraph{Illustration of VCs:} We show (Figure \ref{fig:vcs_general}) that VCs can capture high level, global and some aspects in the scene that can not be shown explicitly or detected by other techniques (\emph{e.g.}, Object detectors as in OSCAR \cite{li2020oscar}).

\begin{table}[!t]
    \small
	\centering	
	\resizebox{0.3\textwidth}{!}{%
	\begin{tabular}	{l |  c  c  c }
		\toprule	 	
	 Method & Val & TestA & TestB  \\
	  \midrule

	  ARN~\cite{Liu_2019_ICCVarn} &  32.78 & 34.35 & 32.13  \\
	  CCL~\cite{ccl}  &  34.29 & 36.91 & 33.56 \\
	  KPRN~\cite{liu2019knowledge} & 35.96 & 35.24 & 36.96  \\
	  DTMR~\cite{sun2021discriminative} & 39.18 & 40.01 & 38.08  \\
	  ALBEF~\cite{li2021albef} & 58.46 & 65.89 & 46.25  \\
	  \midrule
	  ALBEF$^*$ &  57.00& 65.51 & 44.24 \\
	  ViCHA &  57.67 & 66.57 & 45.63 \\
	  \midrule
	  ViCHA$^\dagger$ & 56.40   & 65.93  & 45.69  \\
	   
	\bottomrule
	\end{tabular}
 	}
 \vspace{2ex}
	\caption
	{
	\small	
		Comparison with SOTA on Weakly Supervised Visual Grounding RefCOCO+.
	}
	\label{tab:grounding}

\end{table}



\section{Conclusion}
We propose a new efficient VLP approach centered on 3 main components; stronger Vision-Language pre-alignment through hierarchical contrastive objective, self supervision via masked image modeling based on MAE, and a new Visual Concepts injection and extraction technique. The approach outperforms state of the models trained on the same setup (+3.6\% F30K TR and +1.04 Acc. VQA compared to ALBEF), as well as others trained on much more data (times 4 more data, such as OSCAR and ViLT). Overall, we show that investing in the learning schemes is a very promising approach that helps to exploit more effectively the data, especially at low data regime. We hope that this work will encourage more effort (from a wider range of research laboratories) in this direction, that might lead to have more mature techniques, and then, perhaps, wisely leverage them when going large scale.

\section{Acknowledgments} The authors would like to thank Corentin Dancette and Arthur Douillard for fruitful discussion and Christophe Bouder for technical support. This work was partly supported
by ANR grant VISA DEEP (ANR-20-CHIA-0022), and
HPC resources of IDRIS under the allocation 2022-[AD011013415] and 2022-[A0121012449] made by GENCI.

\clearpage

\appendix
\section{Appendix}
The Appendix is organised as follows; in section \ref{app:tasks}, we give more details about the downstream tasks and how we finetune the model on them, then in section \ref{app:implem}, we elaborate on the implementation details used to train ViCHA. We do more ablation study in section \ref{app:ablation}. We conduct additional experiments on data filtering in section \ref{app:filtering}. We discuss some aspects of the paper in section \ref{app:discussion} and conclude the appendix in section \ref{app:qualitative} by showing some qualitative results on visual grounding, VQA and  some limitations of VCs.

\subsection{Downstream Tasks}
\label{app:tasks}
We mainly follow the evaluation setup of ALBEF \cite{li2021albef}. During finetuning, the image resolution is increased to 380. We use 4 GPUs for Image-Text retrieval and 2 GPUs for other tasks.

\paragraph{Image-Text Retrieval} consists of finding the image that corresponds to a given text or vice-versa. We finetune and evaluate on MSCOCO \cite{lin2014microsoftcoco} with Karpathy split \cite{karpathy2015deep} (113k/5k/5k as train/validation and test set) and Flickr30k \cite{plummer2015flickr30k} (29k/1k/1k). We extract the visual concepts from the training set of these datasets. We train using ITM and ITC objectives and evaluate by selecting top m examples from the global similarity at the output of the dual encoders then re-ranking these examples using the multimodal encoder. We finetune the model for 10 epochs with batch size 64 (16 per GPU), learning rate that decays from $1e-5$ to $1e-6$ using cosine annealing scheduler.

\paragraph{Visual Question Answering (VQA)} consists of answering a question based on an image. We finetune on VQA v2 dataset \cite{goyal2017makingvqav2} that contains images from COCO and is split as 83k/41k/81k images for train/validation/test. The visual concepts are extracted from the COCO training set. We consider the task as a generation problem constrained to the set of possible answers. We add a 6-layer transformer decoder initialized by the pretrained multimodal encoder, which generates the answer based on the output of the multimodal encoder. We train on the train and validation set and did not include additional question and answers from Visual Genome as done in other work \cite{li2021albef, chen2020uniter}, as it slows significantly the training. We finetune for 8 epochs, batch size 32 (16 per GPU), learning rate $2e-5$ warmed up to $1e-5$ and decayed using cosine scheduler to $1e-6$.

\paragraph{Visual Entailment (SNLI-VE) \cite{xie2019visualsnli}} is a visual reasoning task where given an image and a sentence the model should predict if the sentence is a contradiction, neutral or entailment to the image. We train on SNLI-VE \cite{xie2019visualsnli} dataset, which is based on SNLI \cite{bowman2015largesnli} and Flickr30K and consists of 30K images for training and 1k for evaluation and test. The visual concepts are extracted from the training set of Flickr30k. We regard this task as a three-way classification and we add a classification head on top of the [CLS] token of the mulltimodal encoder, consisting of 2 fully connected layers with ReLU activation in between. We train for 10 epochs with batch size 32 (16 per GPU) and initial learning rate 2e-5.

\paragraph{Natural Language For Visual Reasoning (NLVR$^2$) \cite{suhr-etal-2019-corpusnlvr}} is another visual reasoning task. The model should predict if a sentence describes the 2 input images. We consider the task as a binary classification task and we replicate each layer in the multimodal encoder so that the encoder can accept 2 images instead of 1, the replicated layers are initialized from the original ones and the linear projection of the keys and values of the cross attention layers are shared between the original and the replicate.  The visual concepts are extracted from the sentences of the NLVR$^2$ training set. The prediction is obtained from a classification head on top of the multimodal encoder's [CLS] token. We pretrain the model on the pretraining corpus for additional 1 epoch on a classification task (text assignment) that consists of assigning the test to either one of the 2 images or non of them \cite{li2021albef}. We finetune the model on \cite{suhr-etal-2019-corpusnlvr} for 10 epochs, batch size 16 (8 per GPU) with initial learning rate 2e-5.

\paragraph{Visual Grounding} Consists of matching a text query to the corresponding image region. Here we evaluate our approach in a weakly supervised manner where we only use the textual query/caption without using bounding boxes. During training, we train with ITC and ITM loss as we do for retrieval, and during inference we use Grad-CAM \cite{selvaraju2017grad} to rank the proposals from \cite{yu2018mattnet}. We evaluate on the RefCOCO+ dataset \cite{yu2016modelingrefcoco+} which contains 140K expression for 20K images collected from the training set of COCO. The model is trained for 5 epochs without random cropping and initial learning rate of $1e-5$. The batch size is set to 32 (16 per GPU).

\subsection{Implementation Details}
\label{app:implem}
Here we describe additional implementation details to train ViCHA.
The visual encoder is ViT-B/16 \cite{dosovitskiy2021anvit} initialized with DeiT pretrained on ImageNet \cite{touvron2021training}, the text encoder is the first 6 layers of BERT-base \cite{devlin2018bert} and the multimodal encoder is the last 6 layers of BERT-base. $E_{vc}$ is the first 2 layers of BERT-base. We extract 15 concepts for each image and we set $p_{vc}$=30\% for VCA. For H-ITC loss, we set $\lambda_{H-ITC}=0.1$ and align the last 6 layers of $E_v$ with all 6 layers of $E_l$. For U-MIM, we use 2-layer transformer encoder with 16 heads and set $\lambda_{MIM}=1$. The hidden dimension is 768 and kept constant across all modules, the embedding dimension is 256. For the momentum model, the momentum is set to 0.995 and the queue size to 65536. The learned temperature is initialized to 0.07. The images are randomly cropped to $256\times256$ during pretraining. We use AdamW \cite{loshchilov2017decoupledadamw} with learning rate $1e-5$ that is warmed up during the first 2K iterations and then decreased using cosine schedule.

\subsection{Ablation Study}
\label{app:ablation}
In this section, we investigate the importance of different design choices. We follow other approaches \cite{dou2021empiricalmeter, yang2022visiontriplecont} and finetune only on Flickr30K and VQA. We compare against ALBEF$^*$ (Baseline) trained on the same setup described in Section \ref{sec:implem_details}. We keep the image resolution as in pretraining (256).


\begin{table*}[!h]
    \small
	\centering	
 \setlength\tabcolsep{4pt}
	\resizebox{0.7\textwidth}{!}{%
	\begin{tabular}	{ c  c |  c  c  c  c  c  c  c  | c }
		\toprule	 	
	  \multirow{3}{*}{U-MIM} & \multirow{3}{*}{M-MIM} &  \multicolumn{7}{c|}{Flickr30K (1K test set)} & VQA \\
	 &  &  \multicolumn{3}{c}{TR}& \multicolumn{3}{c}{IR} & & test-dev\\
	&  & R@1 &R@5&R@10& R@1 &R@5&R@10& RSUM & Acc.\\
	\midrule
    \multicolumn{2}{c|}{Baseline} &  85.8 & 97.4 & 98.5 & 70.2 & 89.9 & 94.0 & 535.8 & 71.1 \\
    \midrule
	 \ding{51} & & 88.6 & 97.6 & 99.2 & 72.3 & 91.1 & 94.8 & 543.6 & 71.8\\
	 & \ding{51} & 86.8 & 96.4 & 98.5 & 71.8 & 90.8 & 94.7 & 539.0 & 71.8 \\
	 \ding{51} & \ding{51} &  87.4 & 97.6 & 99.2 & 71.4 & 90.8 & 94.6 & 541.0 &  71.9 \\
    \bottomrule

	\end{tabular}
	}
	\vspace{2ex}
	\caption
	{
	\small	
		Ablation study on MIM objective.
	}
    \vspace{-0.5cm}  
	\label{tab:ablation_mim}

\end{table*}

\paragraph{Masked Image Modeling:} Table \ref{tab:ablation_mim} shows the advantage of using MIM objective. Contrary to the results reported in other approaches \cite{dou2021empiricalmeter}, we find that MIM based on the MAE objective brings additional improvements in both cases (U-MIM +7.8\% and M-MIM +3.2\% RSUM). In addition, we found that the unimodal MIM outperforms the multimodal one. This may be due to the interchange between the text and image tokens as query to the multimodal decoder. However, using M-MIM with a decoder that takes the image tokens or both image and text tokens as query could lead to better performance. We tried also to combine both objectives, but we did not notice a significant improvement (+8.8\% RSUM and +0.9\% VQA). Nonetheless, these results reveal the importance of using MIM and SSL objectives for VLP, especially SSL with the unimodal encoders, as it seems improving the unimodal representation lead to better cross-modal alignment.

\begin{table*}[!h]
    \small
	\centering	
 \setlength\tabcolsep{4pt}
	\resizebox{0.6\textwidth}{!}{%
	\begin{tabular}	{c | c | c  c  c  c  c  c | c }
		\toprule	 	
	 VCs & VCs & \multicolumn{6}{c|}{Flickr30K (1K test set)} & VQA \\
	 corpus & corpus & \multicolumn{3}{c}{TR}& \multicolumn{3}{c|}{IR} &  test-dev\\
	& size & R@1 &R@5&R@10& R@1 &R@5&R@10& Acc.\\
	\midrule
    Baseline & - & 85.8 & 97.4 & 98.5 & 70.24 & 89.9 & 94.02 & 71.15 \\
    \midrule
    4OD & 6.1K  & 86.7 & 97.6 & 99.4 &  72.5 & 91.68 & 95.12 & 71.27 \\
    ViCHA & 206.5K & 87.6 & 97.6 & 99.3 &  74.0 & 91.56 & 95.34 & 71.57 \\

    \bottomrule

	\end{tabular}
	}
	\vspace{2ex}
	\caption
	{
	\small	
		Ablation Study: Source of Visual Concepts.
	}
	\label{tab:vc}
\end{table*}

\paragraph{Source of VCs:} How the visual concepts are gathered could affect the performance. This is shown in Table \ref{tab:vc}, where we assess the source of VCs. We found that having a large pool of noisy concepts extracted from captions is better than using a smaller and cleaner pool of concepts extracted from the classes of several object detection datasets (\emph{i.e}, Visual Genome, OpenImage 6, Objects365 and COCO) . However, the number of VCs can be reduced as discussed in Section \ref{app:discussion}.

\begin{table*}[!h]
    \small
	\centering	
 \setlength\tabcolsep{4pt}
	\resizebox{0.6\textwidth}{!}{%
	\begin{tabular}	{c | c  c  c  c  c  c | c }
		\toprule	 	
	 VCs &  \multicolumn{6}{c|}{Flickr30K (1K test set)} & VQA \\
	 Extraction &  \multicolumn{3}{c}{TR}& \multicolumn{3}{c|}{IR} &  test-dev\\
	Method &  R@1 &R@5&R@10& R@1 &R@5&R@10& Acc.\\
	\midrule
    Baseline & 85.8 & 97.4 & 98.5 & 70.24 & 89.9 & 94.02 & 71.15 \\
    \midrule
    VinVL &   88.4 & 98.3 & 99.3 &  73.22 & 92.56 & 95.54 & 72.75 \\
    CLIP &   89.3 & 98.3 & 99.3 &  76.98 & 93.36 & 95.86 & 72.46 \\

    \bottomrule

	\end{tabular}
	}
	\vspace{2ex}
	\caption
	{
	\small	
		Ablation Study: Extraction of Visual Concepts.
	}
	\label{tab:vc_ext}
\end{table*}

\paragraph{Visual Concepts Extraction Method:} Here we show the importance of using our CLIP-based approach for VCs extraction compared to other approaches based on object detectors (\emph{e.g.} VinVL \cite{zhang2021vinvl}). Table \ref{tab:vc_ext} shows that VCs extracted with CLIP give better performance on retrieval and slight degradation on VQA. This might be due to the fact that CLIP excels in retrieval tasks and echos this bias to the extracted VCs.  Importantly, with both approaches, we still get better results than the baseline. This validates that the improvement is not coming only from the large scale pretraining of CLIP.

\begin{table*}[!h]
    \small
	\centering	
 \setlength\tabcolsep{4pt}
	\resizebox{0.6\textwidth}{!}{%
	\begin{tabular}	{c |  c  c  c  c  c  c | c }
		\toprule	 	
	 $E_{vc}$'s & \multicolumn{6}{c|}{Flickr30K (1K test set)} & VQA \\
	 \multirow{2}{*}{\# of layers} & \multicolumn{3}{c}{TR}& \multicolumn{3}{c|}{IR} &  test-dev\\
	& R@1 &R@5&R@10& R@1 &R@5&R@10& Acc.\\
	\midrule
    Baseline &  85.8 & 97.4 & 98.5 & 70.24 & 89.9 & 94.02 & 71.15 \\
    \midrule
    0 &  87.70 & 97.20 & 99.10 &  72.74 & 91.60 & 95.02 & 71.28 \\
    2 &  87.60 & 97.60 & 99.30 &  74.00 & 91.56 & 95.34 & 71.57 \\
    4 &  88.60 & 97.5 & 98.40 &  74.56 & 92.64 & 95.54 & 71.69 \\
    \bottomrule

	\end{tabular}
	}
	\vspace{2ex}
	\caption
	{
	\small	
		Ablation study: size of the Visual Concepts Encoder.
	}
	\label{tab:vc_encoder}
\end{table*}

\paragraph{Size of the Visual Concepts Encoder (VCE):} Having a larger VCE can help to project the VCs to the image patch embedding space and embed VCs more effectively. This is supported by the results in Table \ref{tab:vc_encoder}, showing that increasing the VCE size helps to get better improvements. However, we favor to keep the model simple and use only 2-layers encoder.

\begin{table*}[!h]
    \small
	\centering	
 \setlength\tabcolsep{4pt}
	\resizebox{0.6\textwidth}{!}{%
	\begin{tabular}	{l |  c  c  c  c  c  c | c }
		\toprule	 	
	 \multirow{3}{*}{$p_{vc}$ \%} & \multicolumn{6}{c|}{Flickr30K (1K test set)} & VQA \\
	 & \multicolumn{3}{c}{TR}& \multicolumn{3}{c|}{IR} &  test-dev\\
	& R@1 &R@5&R@10& R@1 &R@5&R@10& Acc.\\
	\midrule

    100 \% (No VCA) &  87.6 & 97.6 & 99.3 & 74.0 & 91.6 & 95.3 & 71.6 \\
    50 \% &  88.5 & 97.5 & 99.0 & 75.6 & 93.3 & 95.4 & 72.2 \\
    30 \% &  89.3 & 98.3 & 99.3 &  76.98 & 93.36 & 95.86 & 72.46 \\
    10 \% &  89.1 & 98.4 & 99.6 &  75.56 & 92.78 & 96.3 & 72.43 \\
    \bottomrule

	\end{tabular}
	}
	\vspace{2ex}
	\caption
	{
	\small	
		Ablation study: the percentage of Visual Concepts Augmentation (VCA) during pretraining.
	}
	\label{tab:vca}
\end{table*}

\paragraph{VCA:} Table \ref{tab:vca} shows the effect of VCA. Reducing the percentage of randomly selected VCs brings significant improvements. This might help the model to avoid relying too much on them by showing different combination at each training step. Interestingly, reducing too much the number of VCs (10 \% or 1 concept/image) impedes the performance. In Table \ref{tab:vca_ft}, we can notice that using VCA only during pretraining is more effective.

\begin{table*}[!h]
    \small
	\centering	
 \setlength\tabcolsep{4pt}
	\resizebox{0.6\textwidth}{!}{%
	\begin{tabular}	{c |  c  c  c  c  c  c | c }
		\toprule	 	
	 \multirow{3}{*}{VCA during FT} & \multicolumn{6}{c|}{Flickr30K (1K test set)} & VQA \\
	 & \multicolumn{3}{c}{TR}& \multicolumn{3}{c|}{IR} &  test-dev\\
	& R@1 &R@5&R@10& R@1 &R@5&R@10& Acc.\\
	\midrule
      \ding{55} & 91.7  & 98.7 & 99.5 & 77.24 & 94.18 & 96.84 & 73.55 \\
      \ding{51} &  91.3 & 98.6 & 99.7 & 77.16 &  94.74 & 96.96 & 73.30 \\

    \bottomrule
	\end{tabular}
	}
	\vspace{2ex}
	\caption
	{
	\small	
		Ablation study: whether VCA is useful during finetuning (FT).
	}
	\label{tab:vca_ft}
\end{table*}

\paragraph{VCs:} We investigate the improtance of VCs during pretraining vs finetuning. Table \ref{tab:vc_ft} shows that using VCs only during pretraining degrades the performace compared to not using them at all. Interestingly, using VCs only during finetuning gives better results than using them also during pretrain. However, we decided to use them also during pretraining  due to the significant boost in performace when VCs are combined with VCA (Table \ref{tab:vca}). Note that, VCA during finetune does not bring additional improvement.

\begin{table*}[!h]
    \small
	\centering	
 \setlength\tabcolsep{4pt}
	\resizebox{0.6\textwidth}{!}{%
	\begin{tabular}	{c c |  c  c  c  c  c  c | c }
		\toprule	 	
	 \multicolumn{2}{c|}{\multirow{2}{*}{VCs}} &  \multicolumn{6}{c|}{Flickr30K (1K test set)} & VQA \\
	 & & \multicolumn{3}{c}{TR}& \multicolumn{3}{c|}{IR} &  test-dev\\
	PT & FT & R@1 &R@5&R@10& R@1 &R@5&R@10& Acc.\\
	\midrule
      \ding{55} & \ding{55} &  85.8 & 97.4 & 98.5 & 70.2 & 89.9 & 94.0 & 71.1 \\
      \ding{55} & \ding{51} &   88.6 & 97.9 & 99.2 & 73.0 &  91.8 & 95.4 &  71.7 \\
      \ding{51} & \ding{55} &  82.0 & 95.3 & 98.2 & 66.2 &  87.6 & 92.5 &  70.3  \\
      \ding{51} & \ding{51} &   87.6 & 97.6 & 99.3 & 74.0 & 91.6 & 95.3 & 71.6 \\
      \midrule
      \multicolumn{9}{l}{\textcolor{gray}{+ VCA ($p=50\%$)}}\\
      \midrule
    \ding{55} & \ding{51} &  88.0 & 97.6& 99.4 & 72.1 & 92.4 & 95.6 & 71.7 \\
    \ding{51} & \ding{51} &  89.3 & 98.3 & 99.3 & 77.0 & 93.4 & 95.9 & 72.5 \\
    \bottomrule
	\end{tabular}
	}
	\vspace{2ex}
	\caption
	{
	\small	
		Ablation study: Visual Concepts (VCs) used during pretraining (PT) and/or finetuning (FT).
	}
	\label{tab:vc_ft}
\end{table*}

\begin{table*}[!h]
    \small
	\centering	
 \setlength\tabcolsep{4pt}
	\resizebox{\textwidth}{!}{%
	\begin{tabular}	{l |  c |  c  c  }
		\toprule	 	
	 Dataset & Corpus size & Examples \\
	\midrule
    COCO & 18K & man, people, street, pizza, laptop, baseball player, tennis court, dogs, rain, bicycles, computer keyboard, airport runway \\
    VG & 78.5K & person, sky, grass, giraffe, airplane, street light, license plate, coffee table, tennis shoes, clothes, sunlight, orange shirt \\ 
    SBU & 110K & house, water, beach, background, office building, kitchen window, netherlands, street signs, eiffel tower, mississippi river, architecture, computer desk \\
    \bottomrule

	\end{tabular}
	}
	\vspace{2ex}
	\caption
	{
	\small	
		Illustration of some VCs corpus for each dataset.
	}
	\label{tab:vc_corpus}

\end{table*}

\subsection{Data Filtering}
\label{app:filtering}

The growing size of image-text datasets scraped from the internet comes also with a growing amount of all kinds of accompanying noise. Even though, the presence of noise is not a major issue for extremely large datasets, the need for efficient data filtering techniques becomes inevitable when considering small to medium scale data regimes. 

Many existing works focus on filtering the dataset from noise based on handcrafted filters \cite{Sharma2018ConceptualCA, changpinyo2021conceptual, desai2021redcaps}. Despite being effective, here we rather focus on selecting the best examples or image-text pairs for the underlying alignment task. 

We propose a simple yet effective technique that selects the best image-text pairs based on the CLIP cosine similarity. This technique, can also cope with noisy and corrupted captions, as these captions have low similarity scores compared to the image. In addition, some captions are hard to align, especially when describing a small region of the input image that can be potentially removed with augmentation techniques (\emph{e.g.} random crop). Moreover, it can filter wrong captions that might be fetched from the internet. Thus, we argue that noise filtering is not enough to have good captions, as they ignore the level of correspondence between the image and the caption. This data selection technique is similar to what have been proposed in \cite{schuhmann2021laion}, however they use a fixed and manually selected threshold, while here we select top $p$ \% pairs. We use the terms data selection and filtering interchangeably.

\paragraph{CLIP-based data selection:} We use CLIP-ViT/B-16 to compute the image features, and the corresponding text encoder to compute the text features of the corresponding caption. We compute the cosine similarity for all pairs then we select the top $p$ \% of the pairs for training. 

\paragraph{Experimental Results:}  Table \ref{tab:data_filter_albef} shows the results of applying this technique to ALBEF$^*$ and ViCHA. We consider COCO as a clean dataset as it is manually annotated for image captioning, and focus on Visual Genome (captions that describe small regions in the image) and SBU (noisy captions). For Visual Genome, we filter only the captions (\emph{e.g.}, 50 $\%$ of the captions for each image). In the case of ALBEF$^*$, we notice that the using all the captions in the Visual Genome (VG) dataset harm the performance, as the results are significantly better when training only with 15 $\%$ and 50 $\%$ of the caption for each image. We obtain comparable performance when training on 50 \% or 70 $\%$ of SBU. In the case of ViCHA, we have a comparable results to training on all data when training with 50 $\%$ VG and 70 $\%$ of SBU.

\begin{table*}[!t]
    \small
	\centering	
 \setlength\tabcolsep{4pt}
	\resizebox{0.8\textwidth}{!}{%
	\begin{tabular}	{l | c c c |  c  c  c  c  c  c | c }
		\toprule	 	
	 \multirow{3}{*}{Model} & \multicolumn{3}{c|}{dataset} & \multicolumn{6}{c|}{Flickr30K (1K test set)} & VQA \\
	  & \multicolumn{3}{c|}{percentage \%} & \multicolumn{3}{c}{TR}& \multicolumn{3}{c|}{IR} &  test-dev\\
	& COCO & VG & SBU & R@1 &R@5&R@10& R@1 &R@5&R@10& Acc.\\
	\midrule

    \multirow{5}{*}{ALBEF$^*$} & 100 & 100 & 100 &  85.8 & 97.4 & 98.5 &  70.24 & 89.9 & 94.02 & 71.04 \\
    & 100 & 50 & 100 &  86.1 & 97.3 & 99.1 &  72.7 & 91.92 & 95.34 & 71.58 \\
    & 100 & 15 & 100 &  86.8 & 97.3 & 99.2 &  71.94 & 91.34 & 95.02 & 71.17 \\
    & 100 & 50 & 50 &  88.5 & 97.6 & 99.1 &  72.38 & 91.52 & 95.12 & 71.38 \\
    & 100 & 50 & 70 &  88.5 & 97.6 & 99.1 &  72.16 & 91.78 & 95.3 & 71.55 \\
    \midrule
    \multirow{4}{*}{ViCHA} & 100 & 100 & 100 & 89.7   & 98.6  & 99.8  &  75.78  & 93.46  & 96.54  & 72.55  \\
    & 100 & 50 & 100 & 89.4   & 98.1  & 99.7  & 76.3   & 93.6  & 96.6  & 72.36  \\
    & 100 & 50 & 70 &  88.2  & 98.1  & 99.5  & 75.66   & 93.22  & 96.26  & 72.41  \\

    \bottomrule

	\end{tabular}
	}
	\vspace{2ex}
	\caption
	{
	\small	
		To see later if we include all these results (with 15\% of vg, ViCHA seems useless) CLIP-based data selection/filtering results: we ablate the choice of the selection percentage for each dataset.
	}
	\label{tab:data_filter_albef}
	
\end{table*}



\subsection{Discussion} 
\label{app:discussion}

\paragraph{Visual Concepts:}  We show (in the main paper) that VCs can capture high level, global and some aspects in the scene that can not be shown explicitly or detected by other techniques (\emph{e.g.}, Object detectors as in OSCAR \cite{li2020oscar}). 

\begin{figure}
    \centering
    \subfloat[\tiny{\textbf{Dataset}: COCO \\
    \textbf{Caption}: 'a woman riding a bicycle by two buildings'. \\
    \textbf{VCs}: 'bicycle riding', 'bicycle ride', 'bicycle rides', 'bicycle rack', 'bicycle basket', 'bike carriage', 'bicyclist', 'delivery bicycles', 'bicycle cart', 'biking', 'bicyclist rides', 'bicyclists', 'bicycle rider', 'bike riding', 'bicycler'}]{
        \includegraphics[width=0.45\textwidth, height=0.4\textwidth]{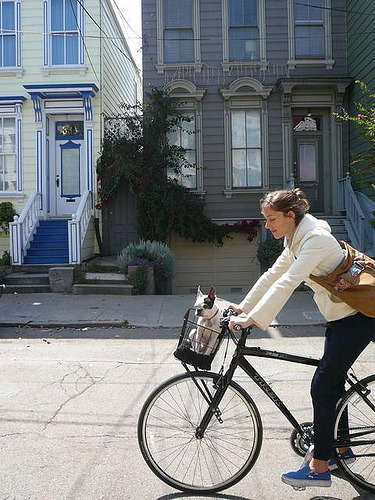}
    }\hfill
    \subfloat[\tiny{\textbf{Dataset}: COCO \\\textbf{Caption}: 'A black and white cow and a man are in the water.'\\
    \textbf{VCs}: 'bear splashes', 'dog swimming', 'bear swimming', 'water buffalo', 'animal swimming', 'holstein cow', 'border collies', 'bear diving', 'elephant swimming', 'calf', 'collie dog', 'border collie', 'retriever dog', 'bull dog', 'splash'}]{
        \includegraphics[width=0.45\textwidth, height=0.4\textwidth]{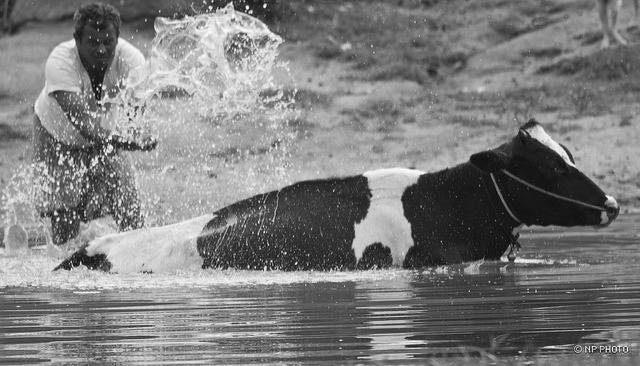}
    }\\
    \subfloat[\tiny{\textbf{Dataset}: COCO\\ \textbf{Caption}: 'A computer is sitting on a computer desk on the far side of the room.'\\
    \textbf{VCs}: 'type bedroom', 'bedroom space', 'home bedroom', 'room combination', 'room arrangement', 'house bedroom', 'bedroom furniture', 'bedroom area', 'studio apartment', 'living someones room', 'someones bedroom', 'bedroom scene', 'bedroom setup', 'college dorm', 'room scene'}]{
        \includegraphics[width=0.45\textwidth, height=0.4\textwidth]{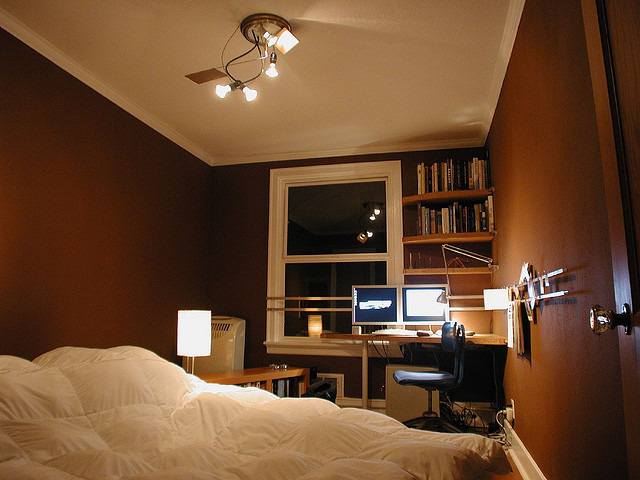}
    }\hfill
    \subfloat[\tiny{\textbf{Dataset}: SBU \\ \textbf{Caption}: 'Kevin is the cutest boy in the world.  He BAKED Chris this cake and decorated it, penis and all!'\\
    \textbf{VCs}: '4.1.2011from', '7/12/2009 view', 'rpop-2008', 'rpop-2007', 'birthday candles', '---cake', 'cake candles', 'trick candles', 'birthday candle', 'guitar cake', '2009.compliments', 'hat cake', '2/21/2010', 'boy baby cake', 'may,09'}]{
        \includegraphics[width=0.45\textwidth, height=0.4\textwidth]{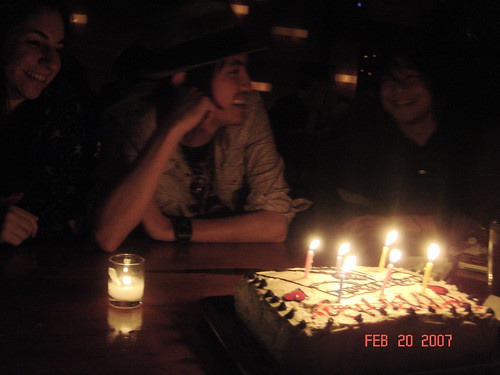}
    }\\
    \vspace{0.2cm}
    \caption{Illustration of some limitations of VCs; (a) redundancy, (b) wrong VCs, (c) lacking local VCs and (d) noisy VCs.}
    \label{fig:vcs_limmit}
\end{figure}

However, our approach is far from being perfect. We show some limitations in Figure \ref{fig:vcs_limmit}, where the VCs can be  (a) redundant, (b) wrong, due to some biases in the dataset, (c) lacking the ability of capture local objects and (d) inherit the noise from the captions. 

There are several reasons for these limitations; (a) The poor (naive) text/captions filtering techniques that we adopt which keep a lot of redundant (\emph{e.g.} plural vs singular, prefixes/sufixes...) and noisy concepts, (b) the retrieval model (CLIP) that is trained for global representation alignment (matching class tokens) which hinders the ability to grasp local concepts. (c) the biases that can be included in the unknown training dataset of CLIP, which cause the model to capture wrong/hallucinated concepts, and finally (d) the limitation of the Scene Graph Parsing technique that we use.

These limitations can be addressed for example by; using some advanced text filtering techniques and eventually a model trained for more finegrained alignment on large unbiased dataset.

\paragraph{Model size:} Another aspect that can help in low data regime, and we little investigate in this work (\emph{e.g.} Table \ref{tab:vc_encoder}), is the size of the model. Recent work \cite{zhai2022scaling, hu2022scaling, lewkowycz2022solving, kaplan2020scaling} show that bigger models are more data/sample efficient. However, in this work we put forward model simplicity and keep this study for future work.

\paragraph{Comparability of existing methods} Despite being common to compare different VLP models, a lot of these models are not comparable due to different design choices, such as; 
\begin{itemize}
    \item The model: size such as base vs large for transformers, single \cite{li2021albef} vs double \cite{dou2021empiricalmeter} multimodal decoders. Different unimodal encoders and initialization; Roberta \cite{dou2021empiricalmeter} vs BERT \cite{li2021albef}, Swin \cite{dou2022coarse} vs CLIP \cite{shen2022how, dou2021empiricalmeter} vs ViT \cite{li2021albef}.
    \item Pretraining dataset: number of images (200K vs 4M vs 14M...), number of image-caption pairs (\emph{e.g.} for 4M images: 5M \cite{li2021albef, yang2022visiontriplecont, duan2022multi} vs 10M \cite{kim2021vilt, dou2021empiricalmeter}). The type/level of annotations used (\emph{e.g.}, bounding boxes \cite{zeng2021multixvlm, dou2022coarse, li2022mvp}). 
    \item Number of epochs: \emph{e.g.}, 10 \cite{dou2021empiricalmeter} vs 30 \cite{li2021albef, yang2022visiontriplecont, duan2022multi}.
    \item Batch size: which is important especially with contrastive learning (\emph{e.g.}, 4096 \cite{kim2021vilt, dou2021empiricalmeter} vs 512 \cite{li2021albef, yang2022visiontriplecont}).
\end{itemize}

For this reason we argue that the best way to assess a given method is to re-implenment or re-train other models with the same setup as the proposed one.

\subsection{Qualitative Results}
\label{app:qualitative}
\begin{figure}[h]
    \centering
    \includegraphics[width=\linewidth]{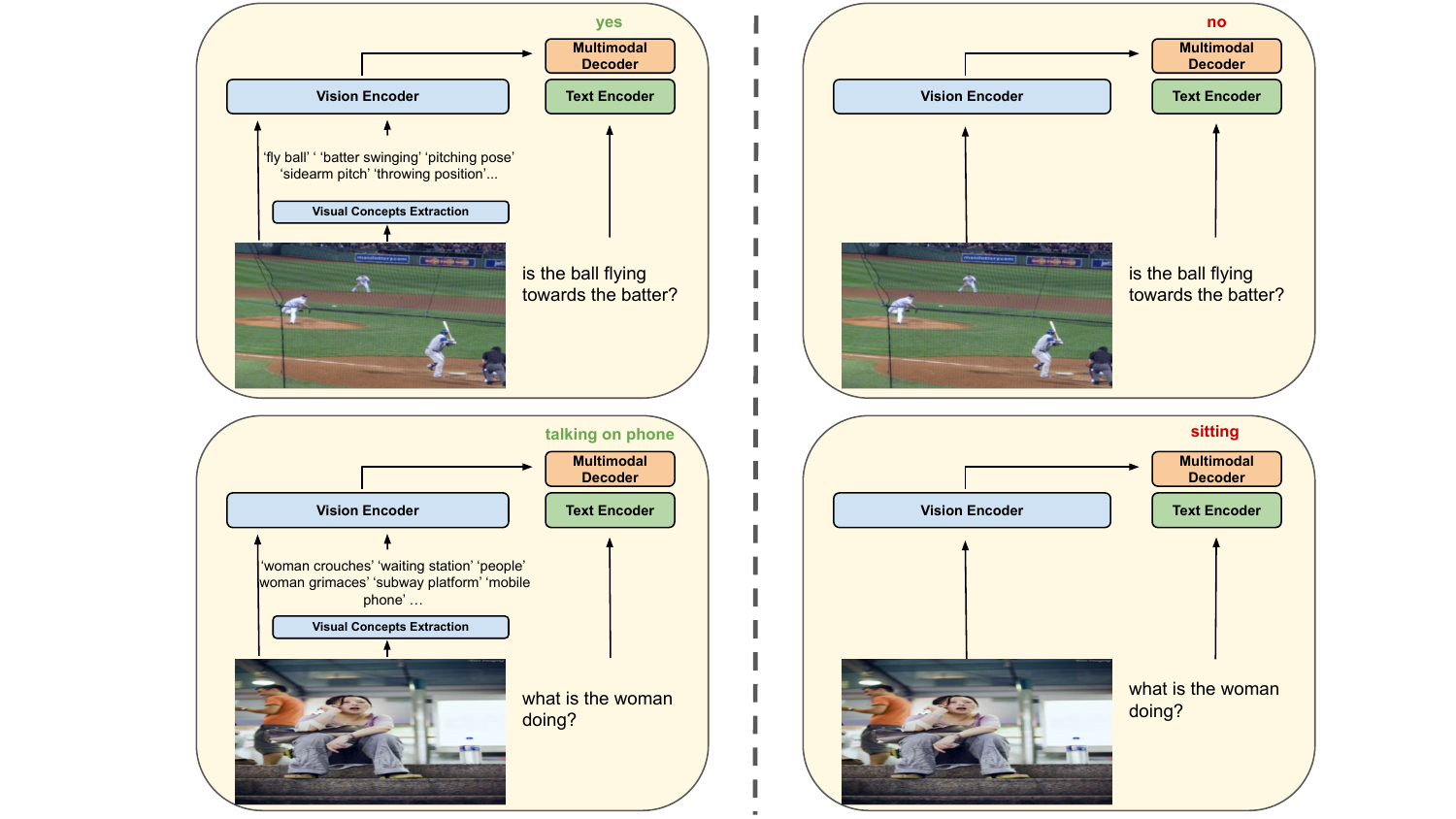}
    \vspace{0.1cm}
    \caption{\small{VQA results: comparison with and without using VCs}}
    \label{fig:vqa_results}
\end{figure}

\begin{figure}
\captionsetup[subfigure]{labelformat=empty}
\captionsetup{position=top}
    \centering
    \subfloat["little"]{
        \includegraphics[width=0.22\textwidth]{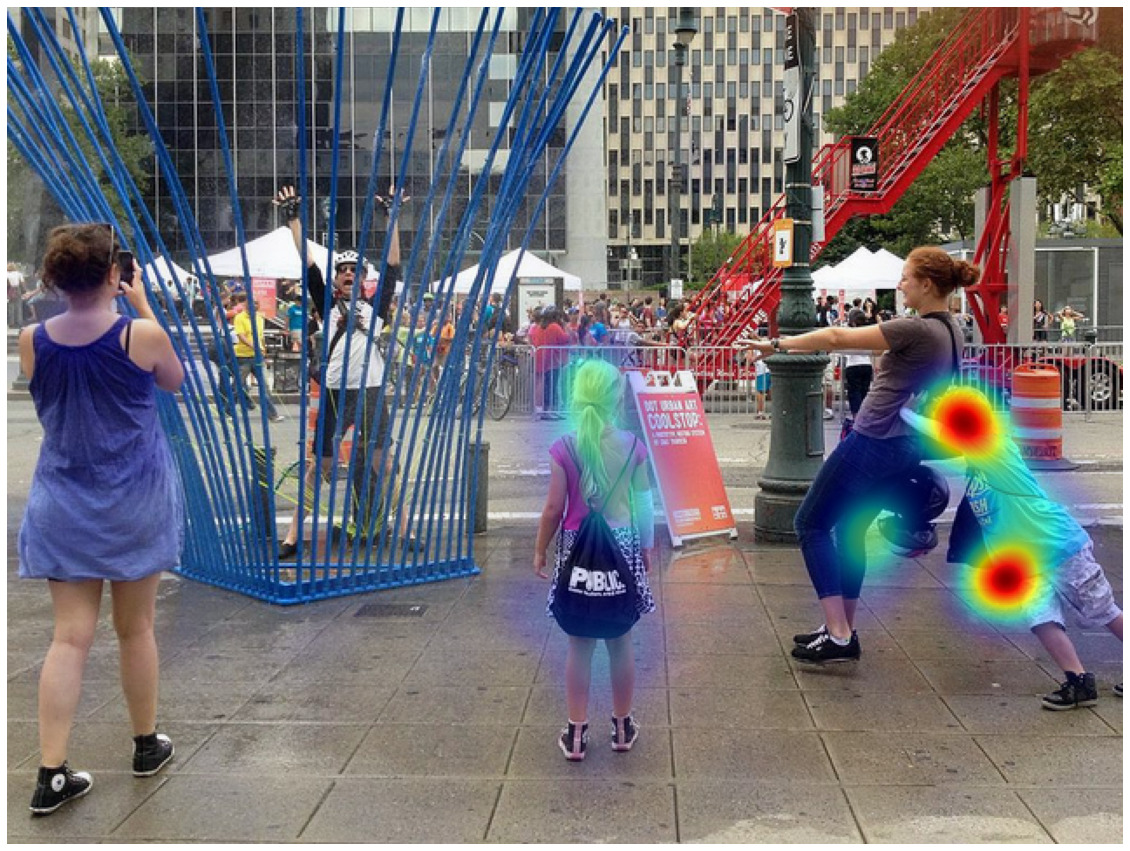}
    }
    \subfloat["pink shirt"]{
        \includegraphics[width=0.22\textwidth]{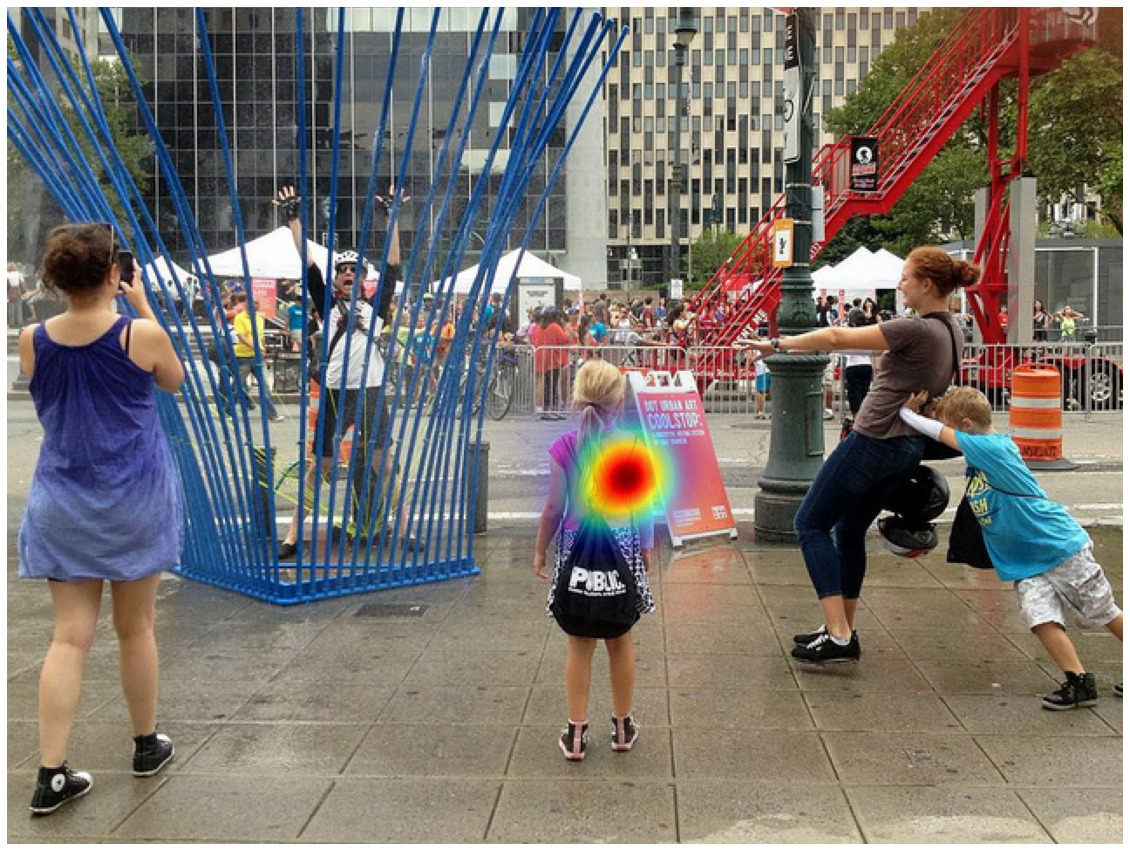}
    }
    \subfloat["standing"]{
        \includegraphics[width=0.22\textwidth]{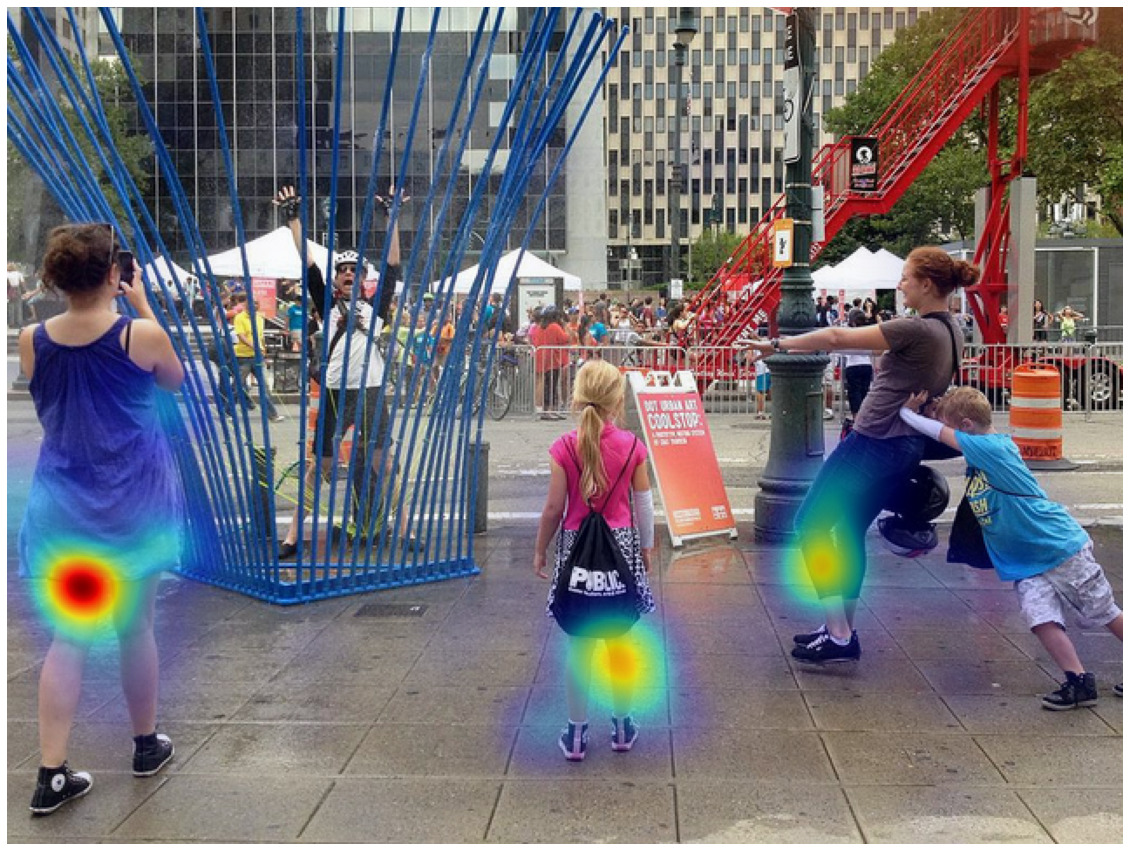}
    }
    \subfloat["sculpture"]{
        \includegraphics[width=0.22\textwidth]{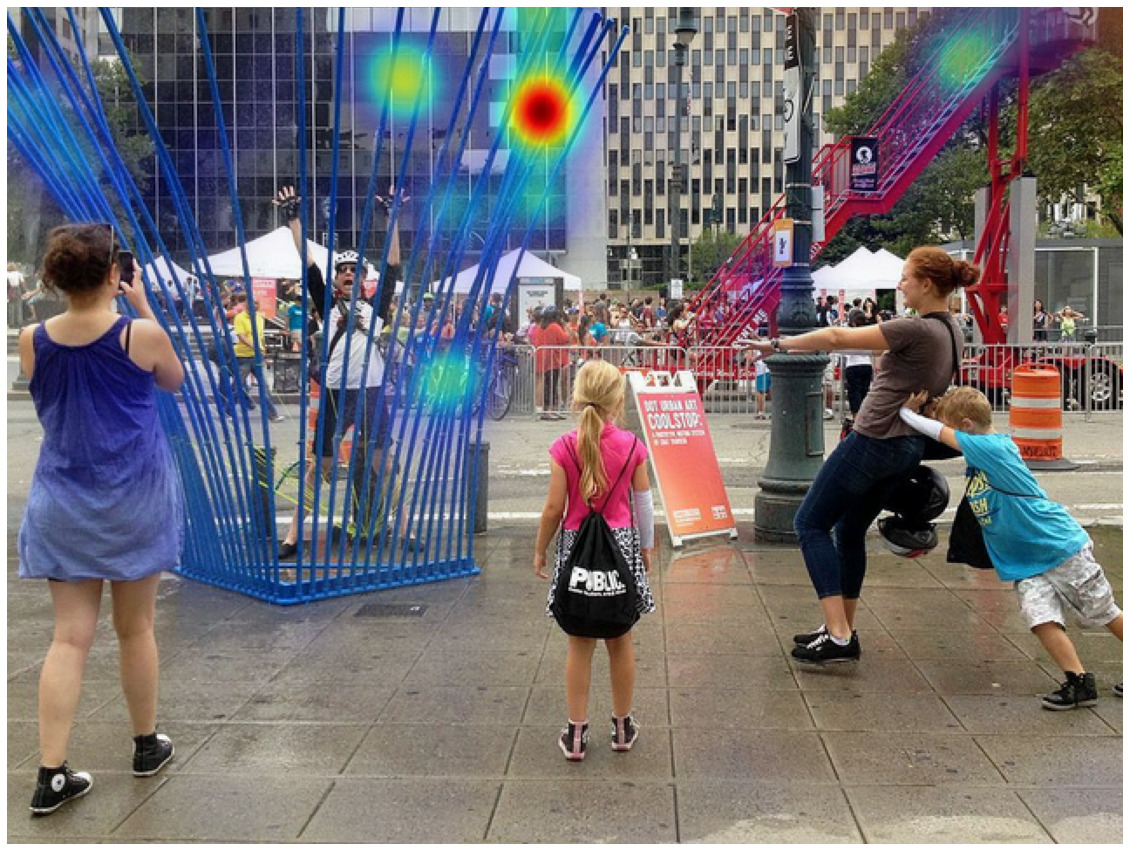}
    }\\\tiny{"A little girl in a pink shirt standing near a blue metal sculpture"}\\
        \subfloat["pot"]{
        \includegraphics[width=0.22\textwidth, height=0.15\textwidth]{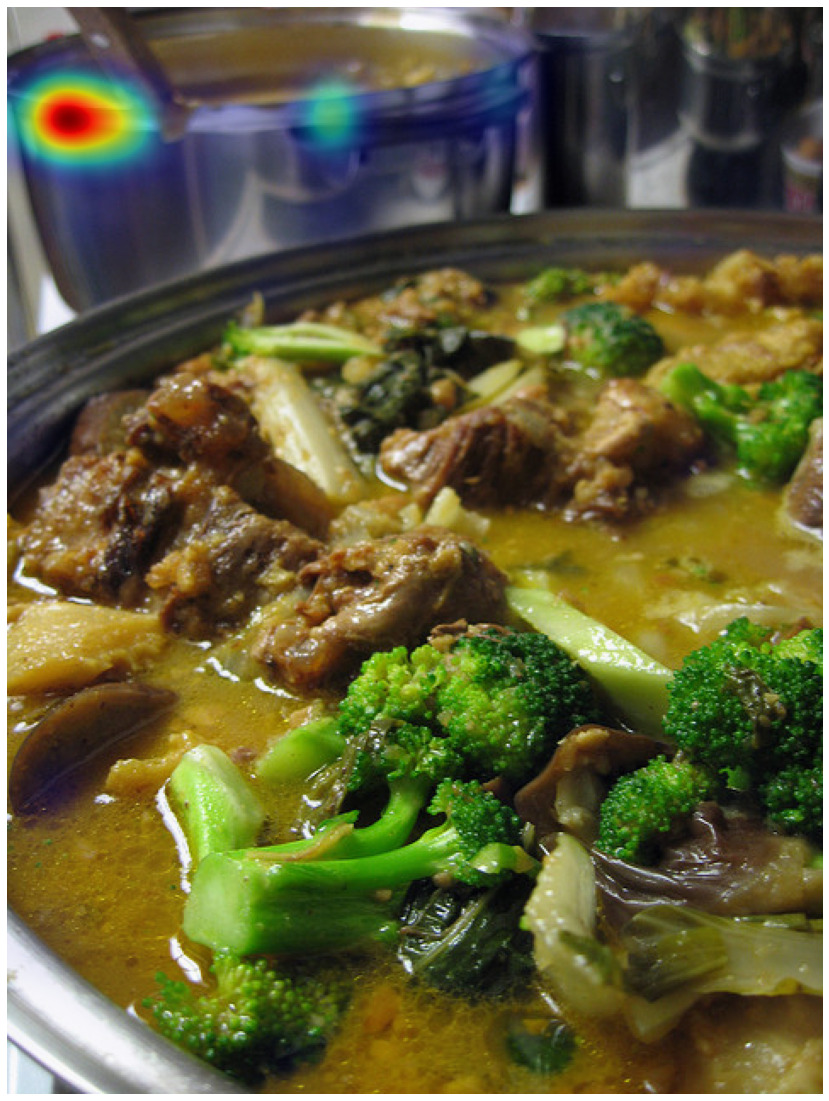}
    }
    \subfloat["beef"]{
        \includegraphics[width=0.22\textwidth, height=0.15\textwidth]{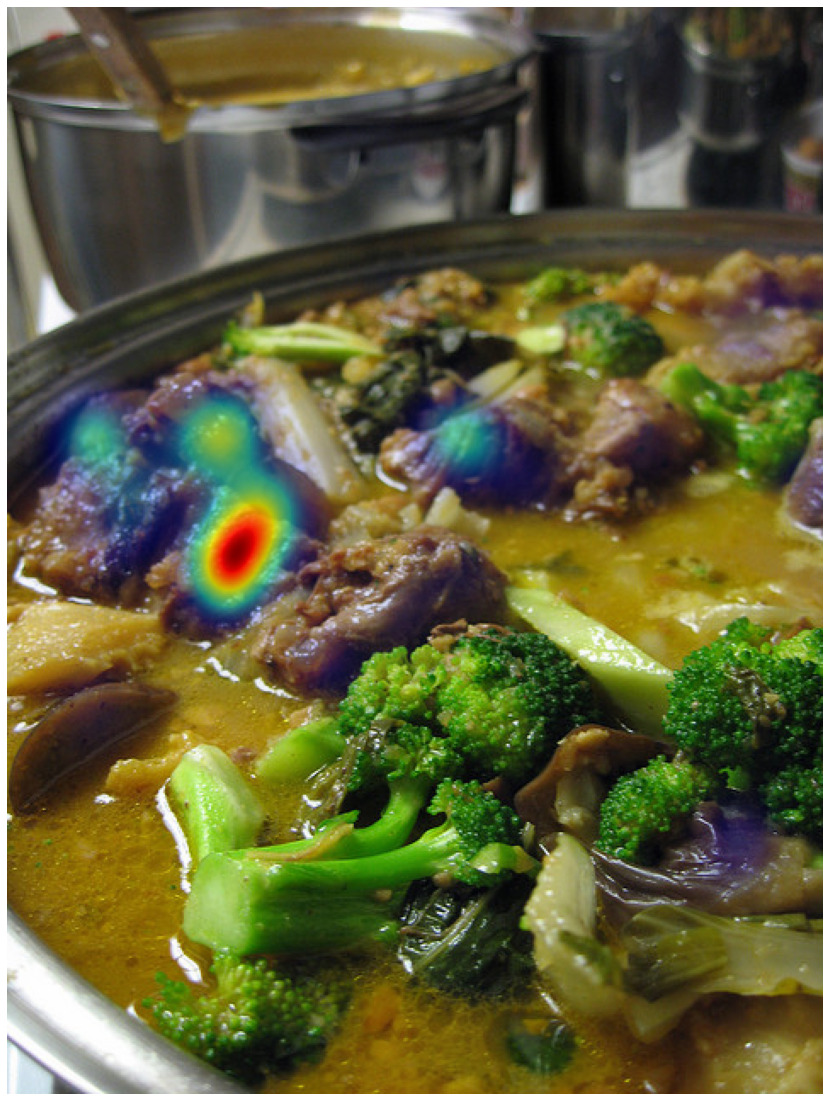}
    }
    \subfloat["brocolli"]{
        \includegraphics[width=0.22\textwidth, height=0.15\textwidth]{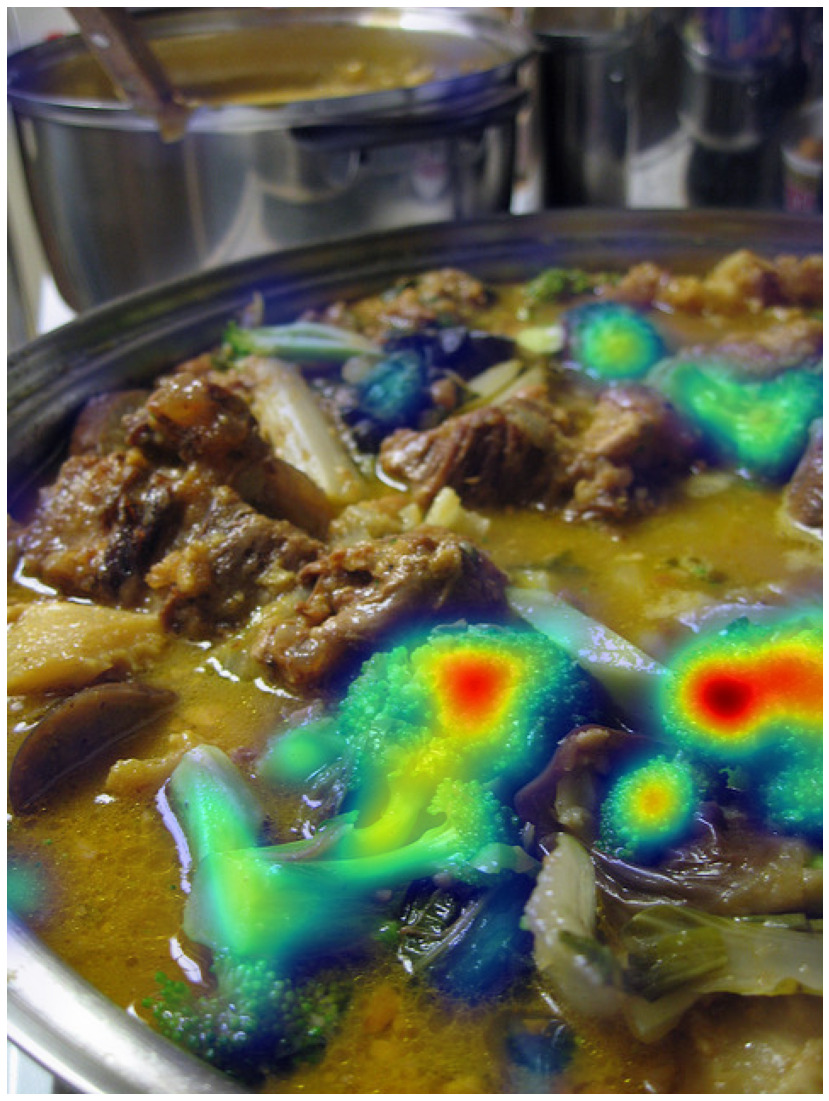}
    }
    \subfloat["stew"]{
        \includegraphics[width=0.22\textwidth, height=0.15\textwidth]{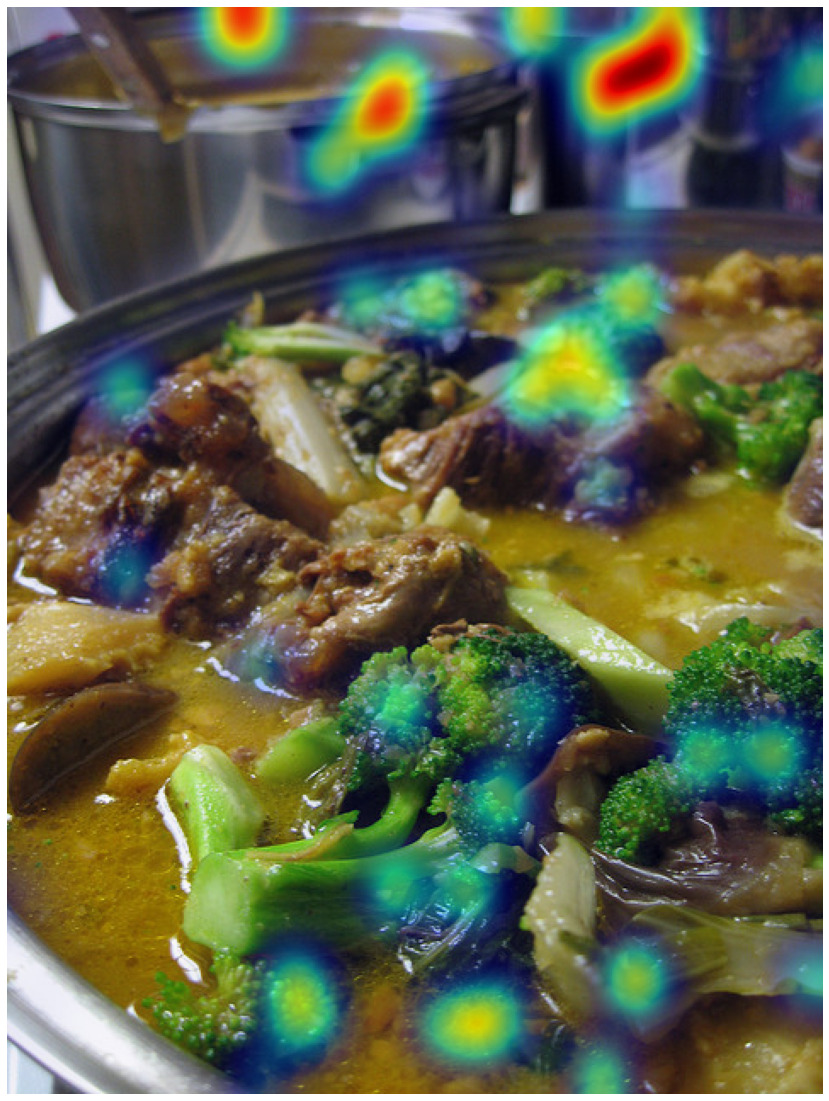}
    }\\\tiny{"A pot full of beef and broccoli stew"}\\
        \subfloat["man"]{
        \includegraphics[width=0.22\textwidth]{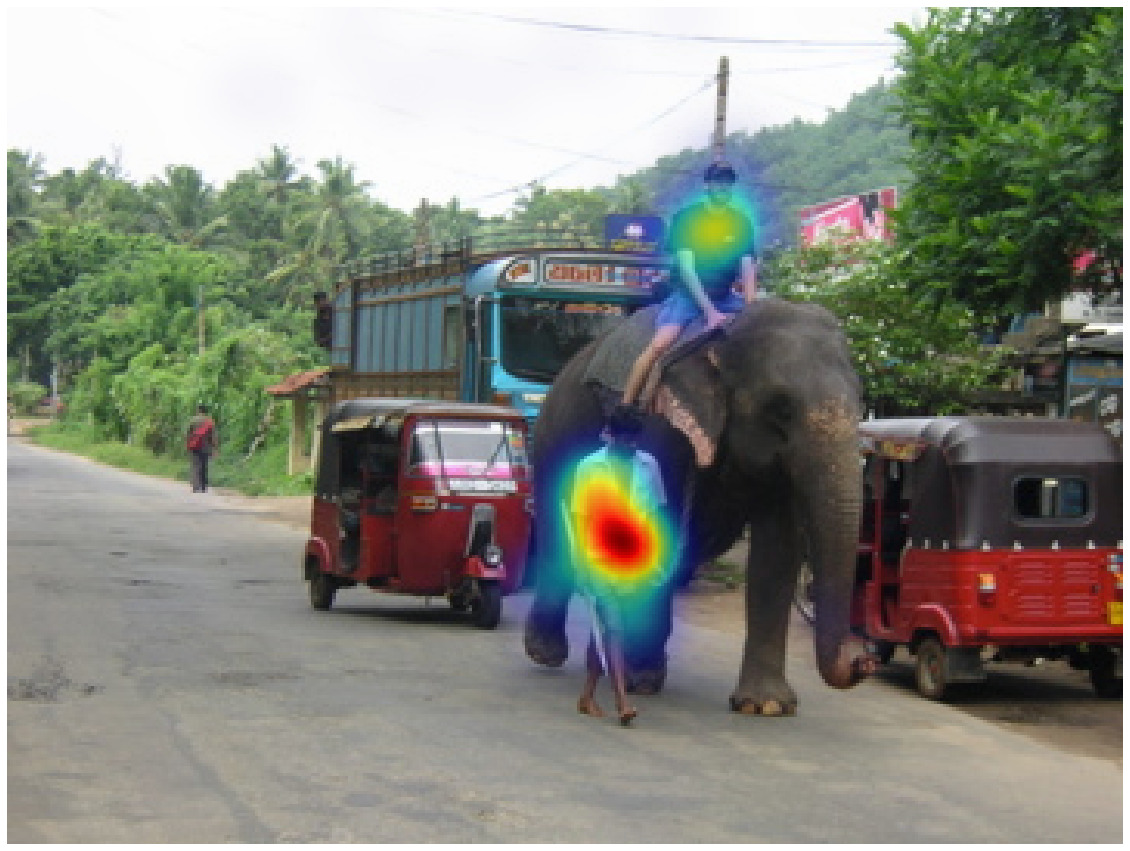}
    }
    \subfloat["black shirt"]{
        \includegraphics[width=0.22\textwidth]{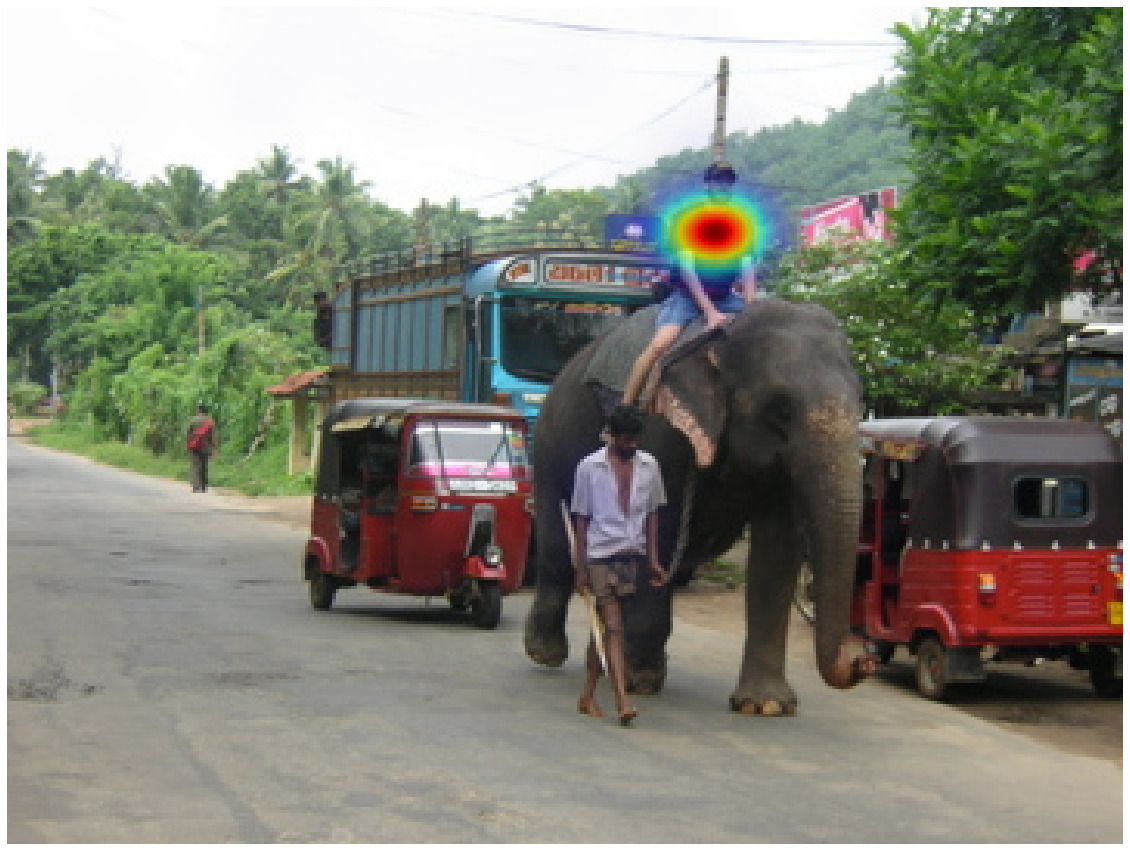}
    }
    \subfloat["elephant"]{
        \includegraphics[width=0.22\textwidth]{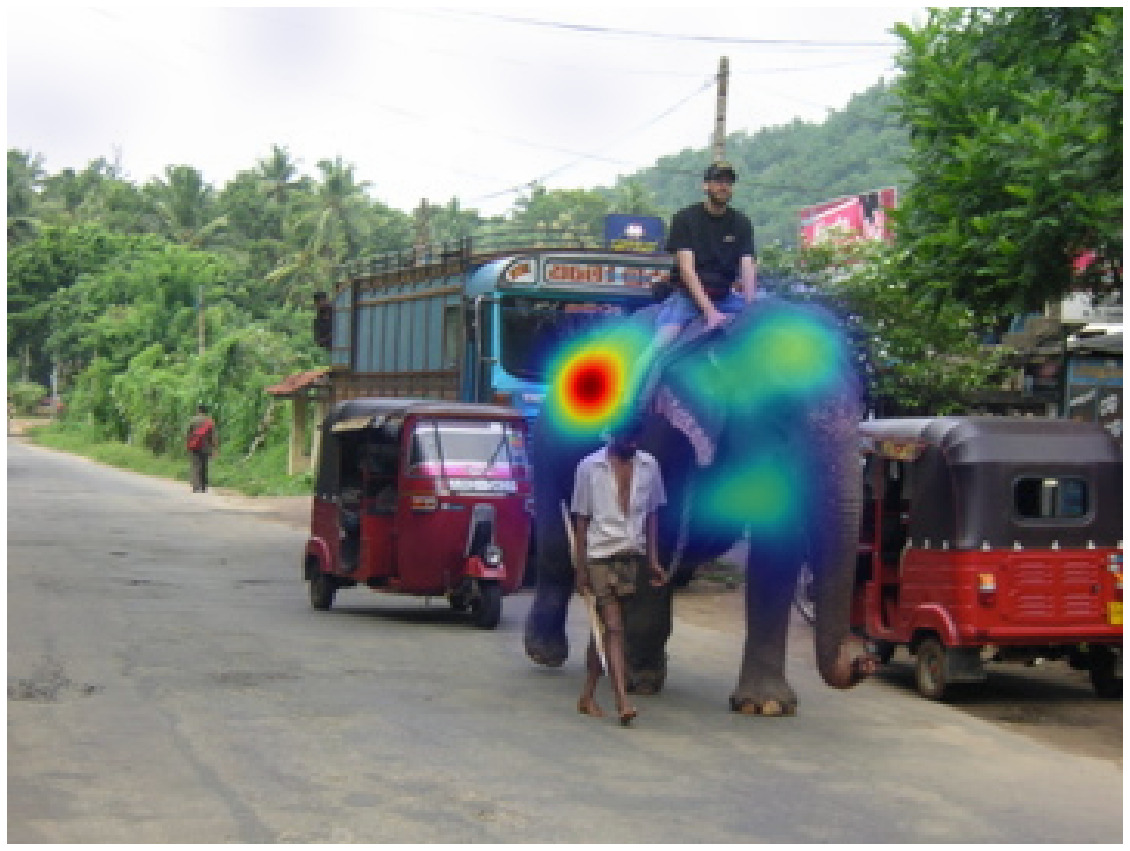}
    }
    \subfloat["walks"]{
        \includegraphics[width=0.22\textwidth]{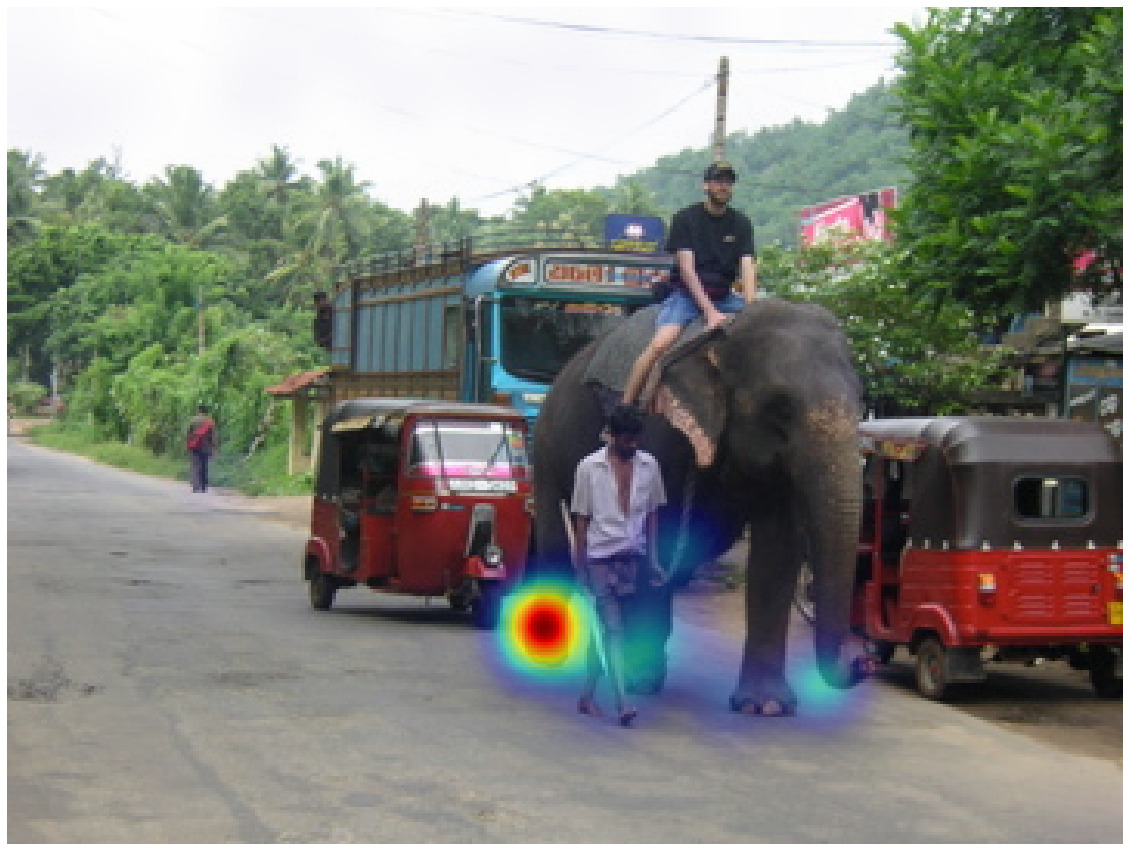}
    }\\\tiny{"A man in a black shirt rides an elephant as a man walks near it down a street"}\\
    \vspace{0.5cm}
    \caption{Visualization of the Grad-CAM corresponding to different words in the caption. The model is finetuned on RefCOCO+ and the images are from COCO test set. The model can accurately localize the objects mentioned in the caption, and also works well with specialized domains, such as food.}
    \label{fig:grounding}
\end{figure}

\clearpage

\bibliographystyle{splncs04nat}
\bibliography{main}
\clearpage
\end{document}